\documentclass[journal]{IEEEtran}
\usepackage{graphicx} 
\usepackage{epstopdf}
\usepackage{amsmath,amsfonts}
\usepackage{algorithmic}
\usepackage{algorithm}
\usepackage{array}
\usepackage[dvipsnames, svgnames, x11names,table,xcdraw]{xcolor}
\usepackage[caption=false,font=normalsize,labelfont=sf,textfont=sf]{subfig}
\usepackage{color}
\usepackage{textcomp}
 \usepackage[table,xcdraw]{xcolor}
\usepackage{stfloats}
\usepackage{url}
\usepackage{verbatim}
\usepackage{graphicx}
\usepackage{booktabs}
\usepackage[table,xcdraw]{xcolor}
\usepackage[normalem]{ulem}
\useunder{\uline}{\ul}{}
\usepackage{graphicx}
\usepackage{amsmath, bm} 
\usepackage{multirow}
\usepackage{float}
\usepackage{cite}

\begin{document}
\setlength{\belowdisplayskip}{4pt}
\title{VDIP-TGV: Blind Image Deconvolution via Variational Deep Image Prior Empowered by Total Generalized Variation}

\author{Tingting Wu$^{1}$, Zhiyan Du$^{1}$, Feng-Lei Fan$^{3}$, Zhi Li$^{2}$, 
Tieyong Zeng$^{3}$

\thanks{ This work was supported in part by the National Key R$\&$D Program of
 China under Grants 2021YFE0203700, NSFC/RGC N$\_$CUHK 415/19, ITF
 MHP/038/20, CRF 8730063, RGC 14300219, 14302920, and 14301121; in
 part by CUHK Direct Grant for Research; in
 part by  the Natural Science Foundation of China (Grant No. 61971234), the Scientific Research Foundation of NUPT (Grant No. NY223008).  (Corresponding author: Tieyong Zeng.)}
\thanks{$^{1}$School of Science,
Nanjing University of Posts and Telecommunications, Nanjing, China (e-mail:
 wutt@njupt.edu.cn; 1221087301@njupt.edu.cn).}
\thanks{$^{2}$Department of Computer Science and Technology, East China Normal University, Shanghai,  China (e-mail:
 zli@cs.ecnu.edu.cn).}
\thanks{$^{3}$Center of Mathematical Artificial Intelligence, Department of Mathematics, The Chinese University of Hong Kong, Shatin, Hong Kong (e-mail: hitfanfenglei@gmail.com;
 zeng@math.cuhk.edu.hk).}}

\markboth{submitted to IEEE Transactions on Circuits and Systems for Video Technology, Vol. XX, No. XX, XX 2023}%
{Shell \MakeLowercase{\textit{et al.}}: A Sample Article Using IEEEtran.cls for IEEE Journals}


\maketitle

\begin{abstract}
Recovering clear images from blurry ones with an unknown blur kernel is a challenging problem. Traditional deblurring methods rely on manually designed image priors, while deep learning-based methods often need a large number of well-curated data. Deep image prior (DIP) proposes to use the deep network as a regularizer for a single image rather than as a supervised model, which achieves encouraging results in the nonblind deblurring problem. However, since the relationship between images and the network architectures is unclear, it is hard to find a suitable architecture to provide sufficient constraints on the estimated blur kernels and clean images. Also, DIP uses the sparse maximum a posteriori (MAP), which is insufficient to enforce the selection of the recovery image. Recently, variational deep image prior (VDIP) was proposed to impose constraints on both blur kernels and recovery images and take the standard deviation of the image into account during the optimization process by the variational principle. However, we empirically find that VDIP struggles with processing image details and tends to generate suboptimal results when the blur kernel is large. Therefore, we combine total generalized variational (TGV) regularization with VDIP in this paper to overcome these shortcomings of VDIP. TGV is a flexible regularization that utilizes the characteristics of partial derivatives of varying orders to regularize images at different scales, reducing oil painting artifacts while maintaining sharp edges. The proposed VDIP-TGV effectively recovers image edges and details by supplementing extra gradient information through TGV. Additionally, this model is solved by the alternating direction method of multipliers (ADMM), which effectively combines traditional algorithms and deep learning methods. Experiments show that our proposed VDIP-TGV surpasses various state-of-the-art models quantitatively and qualitatively.
\end{abstract}


\IEEEpeerreviewmaketitle

\section{Introduction}
\IEEEPARstart{A}{s} the imaging technology becomes increasingly prevalent, high-fidelity images as a vital medium for data transmission are increasingly important \cite{clinical}. However, in real-world applications, image blur often occurs due to factors such as object motion and disparities in optical systems. This blurring not only hampers subjective visual perception but also has a detrimental effect on downstream computer vision tasks. To solve this problem, image deblurring stands out as a crucial task within the field of image processing.

Mathematically, the blurring process of a clean image $u$ can be characterized as 
 \begin{equation}
  \begin{small}
     s=k\otimes u+\varepsilon,
 \label{eqn:mathematical_process} 
  \end{small}
 \end{equation}
 where the blur kernel $k$ usually refers to a two-dimensional image whose size is smaller than that of the clear image $u$, $\otimes$ is a two-dimensional convolution that combines a clean image $u$ with a blur kernel $k$, and $\varepsilon$ is the two-dimensional Gaussian noise. 
The kernel values range from 0 to 1, and after normalization, their sum amounts to 1. The objective of image deblurring is to recover the corresponding clear image $u$ from the blurry image $s$ \cite{fan2020soft,1,2}. 
In practical scenarios, the blur kernel is often unknown or difficult to obtain, introducing two unknown variables into Eq. \eqref{eqn:mathematical_process}. This significantly complicates the problem. As such, in recent years, blind image deblurring has been rapidly developed, which can be roughly categorized into traditional methods and deep learning-based methods. 

The traditional blind deblurring methods consist of Maximum a Posteriori (MAP)-based methods \cite{1,2,3,4,5,6,7,8,9,10,cas1} and  Variational Bayes (VB)-based methods \cite{11,13,14,15,17}. MAP-based methods aim to estimate the blur kernel $k$ and the clean image $u$ alternately by maximizing the conditional probability:
\begin{equation}
\begin{small}
 \begin{aligned}
 (u,k)&=\mathop{\arg\max}\limits_{u,k}\log P(u,k|s)\\
    &= \mathop{\arg\max}\limits_{u,k} \log P(s|k,u)+\log P(u)+\log P(k),
    \end{aligned}
     \end{small}
\end{equation}
where $P(u)$ represents the probability distribution of the clean image, $P(k)$ is the probability distribution of the blur kernel, and $P(s|k,u)$ characterizes a fidelity term. The effectiveness of MAP-based deblurring methods relies heavily on accurately extracting salient edges and formulating constraints for the blur kernel. When estimating the blur kernel, the prominent edge is the key to guiding it to move in the right direction, but accurate edges are difficult to achieve. In addition, Fergus \textit{et al.} \cite{13} indicated that when sparse image priors are used as constraints in the deblurring framework of MAP (sparse MAP), all gradients in the image tend to disappear. Levin \textit{et al.} \cite{10} demonstrated that the sparse MAP solely is unsuitable for blind deconvolution, as prior constraints based on MAP are insufficient to enforce the selection of a clear image, often resulting in a trivial solution without deblurring and a delta kernel (the issues of sparse MAP). To solve the above problems, Fergus \textit{et al.} \cite{13} adopted the variable Bayesian (VB) method of Miskin and Mackay \cite{miskin2000ensemble} and used the Gaussian image prior to model natural image statistics. This approach marginalized the image during optimization while estimating the unknown kernel. The issues of the sparse MAP were avoided well by adding the constraint of image deviation. After the success of this method, many VB-based methods have been proposed to solve blind deconvolution problems \cite{15,17,18}.

To enhance the quality of deblurred images, researchers have introduced a broad spectrum of blind deblurring methods based on deep learning. Typically, these methods involve two stages. In the first stage \cite{21,22,23}, deep learning techniques are employed to estimate the blur kernel, which is then used for non-blind image deblurring. The second stage \cite{24,25,26,27,28,cas2,cas3} involves a straightforward end-to-end deep learning approach for image blind deblurring. However, deep learning-based methods may lose efficacy when images contain information not encountered during training, such as unfamiliar blur kernels and features. Therefore, obtaining an image-specific model becomes crucial.

An image-specific model, called Deep Image Prior (DIP) \cite{31}, proposes to use the deep network as a regularizer rather than as a supervised model. Ren \textit{et al.} \cite{dip37} applied DIP to blind deblurring problems. However, since the relationship between images and their corresponding architectures is unclear, it is hard to find a suitable architecture to provide sufficient constraints on the estimated blur kernels and clean images \cite{dip37}. In addition, if we only add sparse priors into DIP to solve the above problems, the earlier-mentioned issues of the sparse MAP will also appear. Therefore, in order to impose constraints while avoiding the problem of the sparse MAP, Dong \textit{et al.} introduced a new method called Variational Deep Image Prior (VDIP) \cite{vdip}. By using the principle of VB, VDIP imposes constraints on both the blur kernel and clean image and takes the standard deviation of the image into account during the optimization process. Although VDIP has made significant progress in blind deblurring, based on experimental results, it still has room for improvement. First, when the size of a blur kernel is large, the blur kernel estimated by VDIP is obviously inaccurate, and the image will be over-sharpened after deblurring. The reason may be that the image variance in the blurry image with a relatively small size contains too little information to estimate the blur kernel with a large size. Second, since VDIP lacks any regular constraints preserving details, the image deblurred by VDIP introduces additional artifacts and loses some edge details.

To solve the above problem, we regularize VDIP with total generalized variation (TGV) \cite{38}. Our model uses not only the variance information of pixels but also TGV constraints to capture more internal structure information or edge information. Therefore, even when the size of the blur kernel is large, our method can capture more information than VDIP solely. Furthermore, the TGV can be seen as a sparse penalty term that optimally balances the first-order distribution derivative and the $k$-th order distribution derivative. It can be regularized selectively at different levels, so that it can remove the stair-casing effect while preserving more details, such as edge information. Thus, the drawback of VDIP not being able to retain more details can also be overcome by introducing TGV.
Next, we find that the optimization of VDIP-TGV is tricky. Directly minimizing the loss function (including the fidelity term and the TGV regularization) like VDIP and DIP-based methods \cite{dip37,34,double-dip} does not work. The value of the loss function required for each network parameter update is calculated based on the output result $u$ of each iteration of the network. However, we find that the TGV in Eq. (\ref{eq:tgv2}) has a variable $q$, which cannot be directly computed by $u$ during the iteration. Considering that the alternating direction method of multipliers (ADMM) \cite{admm1,admm2} is famous for decomposing a complex optimization problem into several simple subproblems, we use the ADMM algorithm to solve the problem. The optimal solution to complex problems is obtained by iteratively solving subproblems.
Our key contributions are twofold:

\begin{itemize}
	
 \item[$\bullet$] We propose VDIP-TGV that exerts an explicit TGV regularization to improve the VDIP-based approach in terms of addressing large-size blur kernels and retaining image details. Our method employs the anisotropic TGV which considers the gradient components to supplement more information in preserving details in deblurred images and estimating the kernels.
 
	\item[$\bullet$]Our experiments show that the proposed VDIP-TGV is superior to VDIP and other state-of-the-art deblurring methods quantitatively and qualitatively on both real and synthetic images.
\end{itemize}


\section{Related Work}

\textbf{Traditional Blind Image Deconvolution}. Traditional methods for blind deblurring can be categorized into two groups: MAP-based methods and VB-based methods. Within the MAP-based methods, several approaches focus on extracting significant edges to accurately estimate blur kernels. For instance, Shan \textit{et al.} \cite{1} and Almeida \textit{et al.} \cite{2} proposed regularized constraints to identify significant edges and improve the accuracy of blur kernel estimation. Cho \textit{et al.} \cite{3} employed a prediction step to select significant edges, while Pan \textit{et al.} \cite{4} developed a model based on total variation denoising to identify image edges suitable for blur kernel estimation. Other MAP-based approaches concentrate on designing constraints for blur kernels. The $L_p (0<p<2)$ norm is commonly used to constrain pixel values, ensuring the sparsity of motion blur kernels. You \textit{et al.} \cite{5} utilized the piecewise smoothness of images and blur kernels, employing the $L_2$ norm to constrain both. Chan \textit{et al.} \cite{6} introduced the total variation model to simultaneously estimate the image and blur kernel in blind deblurring. Pan \textit{et al.} \cite{7} discovered through experimental statistics that the dark channel sparsity of clear images surpasses that of blurry images, leading to a blind deblurring method based on the dark channel prior. Krishnan \textit{et al.} \cite{8} employed the ratio of the $L_1$ norm to the $L_2$ norm of image gradients as a prior for gradient sparsity in blind deblurring. Shao \textit{et al.} \cite{9} used a double $L_0-L_2$ norm as a complex image prior to simultaneously constrain the blur kernel and image. This method reduces gradient artifacts and enhances the accuracy of blur kernel estimation through sparsity improvements. 

VB-based methods have addressed the limitations of sparse MAP approaches by incorporating the standard deviation of images. Within the VB framework, Fergus \textit{et al.} \cite{13} introduced a method for estimating blur kernels. They achieved favorable deblurring results by combining the mixed Gaussian image prior and mixed exponential blur kernel prior. The research conducted by Levin \textit{et al.} \cite{14} demonstrated, both theoretically and experimentally, the advantages of the VB over the MAP. Wipf \textit{et al.} \cite{15} represented the VB as the MAP based on the spatial adaptive sparse image prior. The VB posterior mean estimation reduces the risk of local minima compared to MAP. Theoretical and empirical analyses in \cite{15,17} have consistently confirmed the superiority of VB-based methods over MAP-based ones.

\textbf{Deep learning-based Blind Image Deconvolution}. Deep learning-based methods are divided into two classes. The first class comprises two-step methods: estimating the blur kernel using deep learning techniques and non-blind image deblurring. Sun \textit{et al.} \cite{21} designed a Convolutional Neural Network (CNN)-based model to estimate blur kernels and remove non-uniform motion blur. Chakrabarti \textit{et al.} \cite{22} performed non-blind deblurring in the frequency domain by predicting the Fourier coefficients of the blur kernel. Xu \textit{et al.} \cite{23} proposed an image deconvolution CNN network with a separable convolutional structure for efficient deconvolution.

The second class comprises end-to-end image deblurring methods. Nah \textit{et al.} \cite{24} introduced DeepDeblur, a multi-scale convolutional neural network that achieves end-to-end restoration for blurry images caused by various factors. Tao \textit{et al.}  \cite{25} utilized a simpler network with fewer parameters, called SRN-DeblurNet, which employs a multi-scale recursive loop network to progressively restore blurry images using a coarse-to-fine multi-scale approach. Kupyn \textit{et al.} \cite{26} proposed DeblurGAN-v2, a classical network that introduces a generative adversarial network (GAN) feature pyramid to perform deblurring. Cho \textit{et al.} \cite{27} presented MIMO-UNet, a network with multiple inputs and outputs that incorporates multi-scale features to improve accuracy and reduce complexity. Finally, Wang \textit{et al.} \cite{28} introduced Uformer, a Transformer-based image recovery method that utilizes a novel locally enhanced window Transformer module as a base module to construct a layered encoder-decoder structure.

\textbf{Deep Image Prior}. Ulyanov \textit{et al.} \cite{31} invented deep image prior (DIP), which takes the structure of a randomly initialized network as an image prior to image restoration. The research in GpDIP \cite{32} provided evidence that the DIP converges to a Gaussian process (GP) under specific circumstances. The efficacy of DIP has been further enhanced by incorporating total variational  (TV) regularization  \cite{33} and vector bundle total variational (VBTV) regularization \cite{34}. However, DIP-TV \cite{33}  and DIP-VBTV \cite{34} perform well only when blur kernels are known.
The Double-DIP technique \cite{double-dip} employs two DIP architectures to estimate both the blur kernel and the clean image. To improve the use of DIP for blind image deconvolution, Ren \textit{et al.} \cite{dip37} utilized two unconstrained generative networks (the DIP-FCN architecture instead of the DIP-DIP architecture used in Double-DIP \cite{double-dip}) to generate clean image prior and blur kernel prior, respectively. The experimental results were significantly better than those obtained with Double-DIP \cite{double-dip}. 

\textbf{Total Generalized Variation}. Bredies \textit{et al.} \cite{38} were the first to propose a TGV-based model for the blurry image restoration problem with Gaussian noise. Their model effectively overcomes the step effect and preserves the edge structure information of the image. Since then, many researchers incorporated the TGV regularization \cite{39,40,41,42} into their approaches. Lu \textit{et al.} \cite{39} and Honglu \textit{et al.} \cite{40} introduced novel variants of TGV, namely adaptive weighted TGV and non-smooth TGV, respectively, for image restoration. Zhou \textit{et al.} \cite{41} successfully integrated TGV-based regularization into the image enhancement task. These papers conducted experimental analyses to demonstrate that integrating the TGV-based regularization into their models leads to notable improvements in preserving fine image details.

\section{Preliminaries}\label{sec3}

\subsection{Variational 
Deep Image Prior (VDIP)}\label{sec3.1}

Mathematically, the blurring process is characterized as the following:
\vspace{-0.2cm}
 \begin{equation}
  \begin{small}
     s=k\otimes u+\varepsilon,
 \label{eqn:mathematical_processvdip}  
 \end{small}
 \end{equation}
where $k$ is the blur kernel, $u$ is the clear image, $\otimes$ is a two-dimensional convolution that combines $u$ and $k$, and $\varepsilon$ is the Gaussian noise. 

When DIP \cite{31} was first proposed, it could only solve the non-blind deblurring problem. Such problems are often formulated as optimizing an energy function:
 \begin{equation}
  \begin{small}
\mathop{\min}\limits_{u} E(u;s,k)+R(u),
 \end{small}
 \end{equation}
where $E(u;s,k)$ is a data fidelity term, and $R(u)$ is an image prior. Setting a suitable prior $R(u)$ is challenging. Deep learning-based approaches train the network on a large number of image pairs to capture a prior $R(u)$. Then, according to a surjective $g:\theta \rightarrow u$, non-blind image deblurring problems are rewritten as
 \begin{equation}
  \begin{small}
\mathop{\min}\limits_{\theta}E(u;s,k)+R(g(\theta)).
  \end{small}
  \end{equation}
 
The idea of DIP is that rather than regarding prior information in the image space as a surjective, why not take prior information in the parameter space of a neural network? Accordingly, DIP defines 
$g(\theta)$ as $T_{\theta}(z)$, where 
$T$
 is a deep ConvNet with parameters 
$\theta$, $z$
 is a fixed noise, and $T_{\theta}$
 is initialized randomly. 
Furthermore, in the framework of DIP, as long as the mapping $g$ is well designed, only degraded images are needed in the non-blind deblurring problems. As a result, the problem can be further simplified as
 \begin{equation}
  \begin{small}
\mathop{\min}\limits_{\theta}E(T_{\theta}(z);s,k).
\label{eq:dipe}
 \end{small}
 \end{equation}

DIP only needs the degraded image during training. The input of the network $T$ is a fixed random code $z$, and the output is a degraded image. DIP shows that a deep CNN has an intrinsic ability to learn the uncorrupted image. Therefore, if the training is interrupted in the middle, it will output a restored image; otherwise, the network $T$ can yield the degraded image. In DIP, it is also crucial to cast the appropriate loss function according to the needs of the problems. 

When DIP was applied to solve the problem of image deblurring, the non-blind deblurring problem can be written as $\mathop{\min}\limits_{\theta} {\frac{1}{2}\Vert k \otimes (T_{\theta}(z))-s \Vert}$, and the solution process can be formulated as
\begin{equation}
\begin{small}
	\begin{cases}
\theta=\mathop{\arg\min}\limits_{\theta} {\frac{1}{2}\Vert k \otimes (T_{\theta}(z))-s \Vert}^{2}_{2},      \\
		u=T_{\theta}(z),	
	\end{cases}
   \end{small}
 \label{dip_nonblind}
\end{equation}
where ${\frac{1}{2}\Vert k \otimes (T_{\theta}(z))-s \Vert}^{2}_{2}$ is the loss function for the network to update parameters.
DIP has an edge in real-world circumstances because it needs no training data. Furthermore, the recovery quality achieved by DIP is on par with some very sophisticated techniques that depend on a large amount of well-curated data.

\begin{figure*}[hb]
	\centering
	\setcounter{subfigure}{0}
	\subfloat{
		\begin{minipage}[b]{0.27\linewidth}
			\centering
		\includegraphics[width=5.1cm]{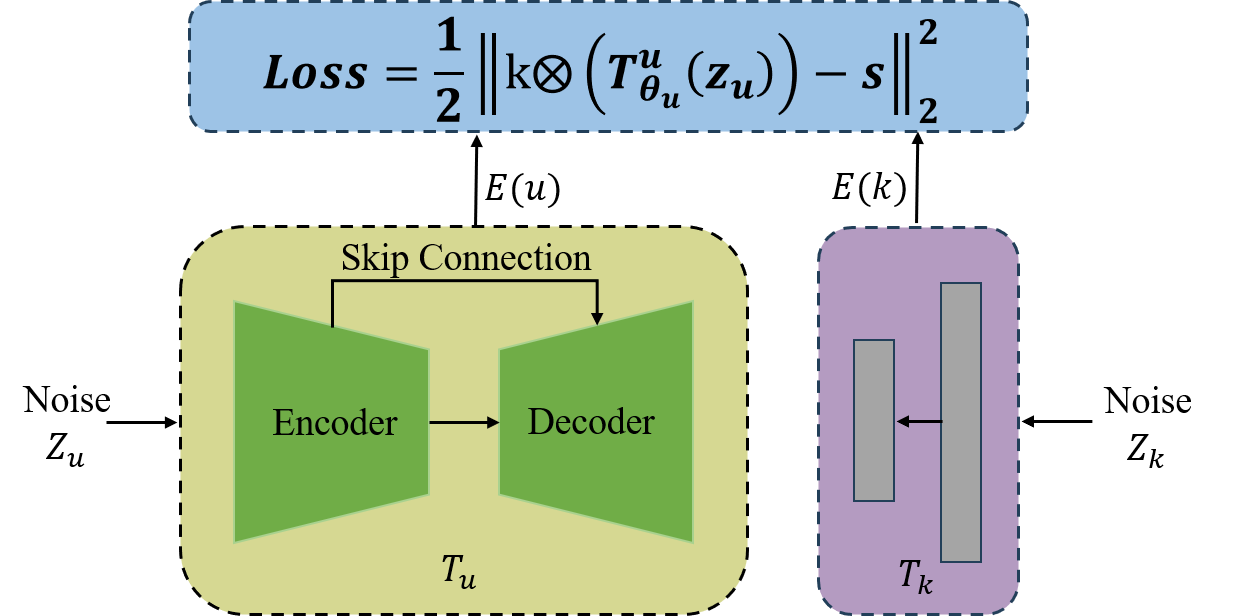}
 \centerline{(a) Overview of the DIP \cite{dip37}.}	
  \end{minipage}
  \begin{minipage}[b]{0.27\linewidth}
			\centering
		\includegraphics[width=5.1cm]{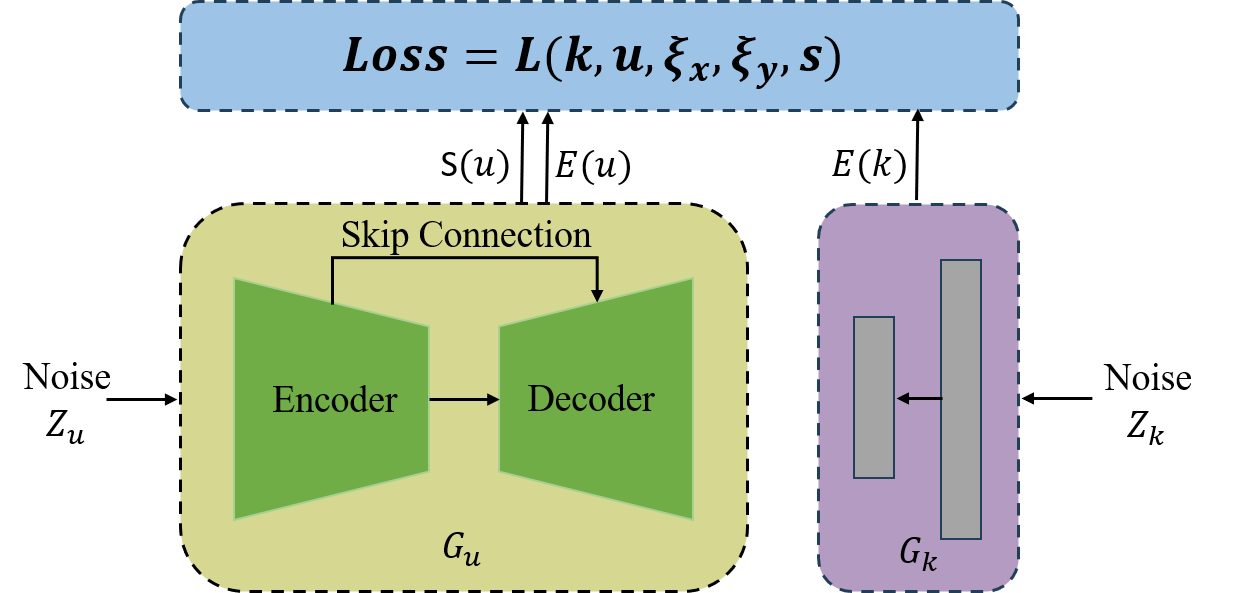}
            \centerline{(b) Overview of the VDIP \cite{vdip}.}
  \end{minipage}
  		\begin{minipage}[b]{0.45\linewidth}
			\centering
		\includegraphics[width=7.5cm]{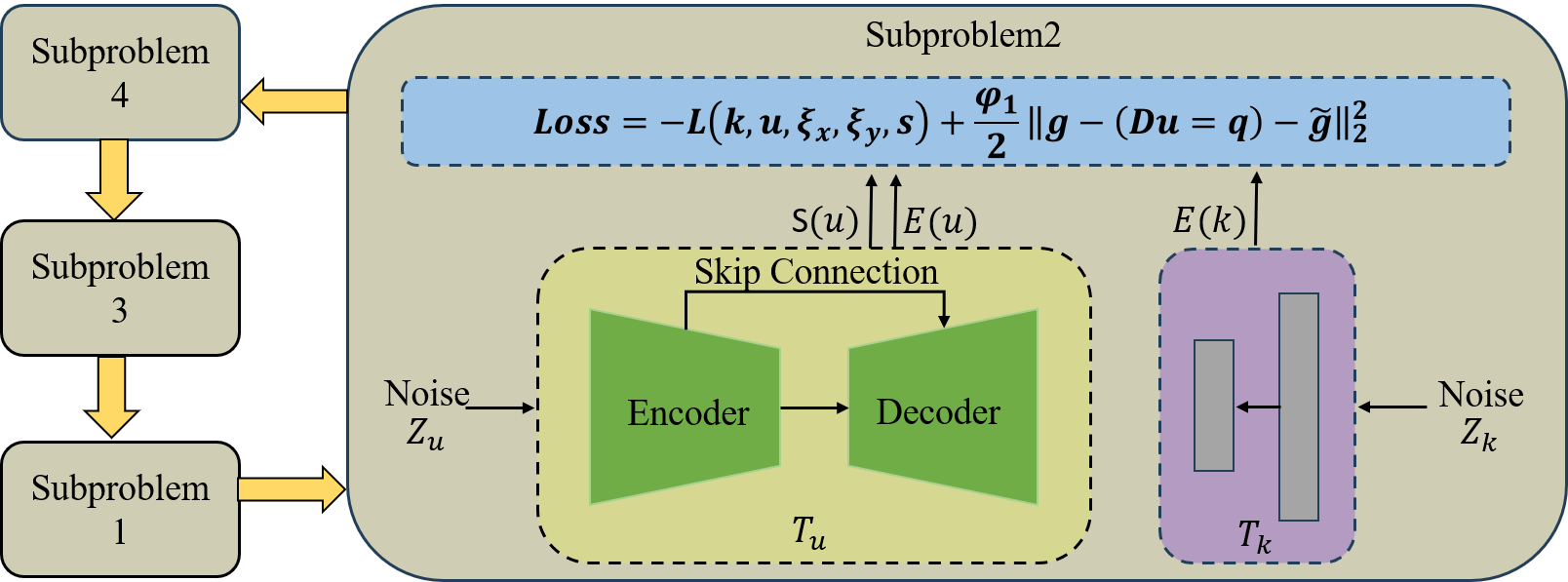}
 \centerline{(c)  Overview of our method.}	
  \end{minipage}
	}
	\caption{The overview diagrams of DIP, VDIP, and the proposed VDIP-TGV. VDIP considers the effect of image variance, thereby bringing extra constraints relative to DIP. Details of the loss function of subgraph (b) can be found in Section \ref{sec3.1}. We introduce TGV regularization on the basis of VDIP. According to the ADMM algorithm, the four sub-problems are solved successively, and the yellow arrow indicates the outer loop. Details of the loss function for subgraph (c) are shown in Section \ref{sec4.2}. }
 \vspace{-3.5mm}
 \label{figoverview}
\end{figure*}

Ren \textit{et al.} \cite{dip37} applied DIP for blind deblurring, and the problem in Eq. (\ref{eq:dipe}) can be formulated as
\begin{equation}
\begin{small}
\begin{aligned}
&\mathop{\arg\max}\limits_{u,k,\theta_u,\theta_k} P(u,k,\theta_u,\theta_k|s) \\
=&\mathop{\arg\max}\limits_{u,k,\theta_u,\theta_k}P(s|u,k)P(u|\theta_u)P(k|\theta_k)P(\theta_{u})P(\theta_{k}),\\
\end{aligned}
\end{small}
\end{equation}
where $P(\theta_u)$ and $P(\theta_k)$ refer to the priors associated with $\theta_u$ (parameters of the image generator $T^u$) and $\theta_k$ (parameters of the kernel generator $T^k$).  The network structure of the image generator $T^u$ is consistent with Eq. (\ref{dip_nonblind}), and the blur kernel generator $T^k$ is composed of fully connected layers. When DIP is used for blind deblurring, the objective function can be expressed by the following formula:
  \begin{equation}
    \begin{small}
	\begin{cases}
 \theta_u, \theta_k=\mathop{\arg\min}\limits_{\theta_u,\theta_k} {\frac{1}{2}\Vert k \otimes (u)-s \Vert}^{2}_{2},      \\
        u=T^u_{\theta_u}(z_u), 
        k=T^k_{\theta_k}(z_k),	
	\end{cases}
 \label{dip_blind}
   \end{small}
  \end{equation}
where ${\frac{1}{2}\Vert k \otimes u-s \Vert}^{2}_{2}$ is the loss function, $z_u$ and $z_k$ are the inputs to the two generators respectively, both in the form of random noise. A frame diagram of DIP for image blind deconvolution can be seen in Fig. \ref{figoverview}. When addressing  image blind deconvolution problems, the DIP assumes constant values for $P(\theta_u)$ and $P(\theta_k)$, which means that DIP does not impose any constraints on the generated images and kernels. 

To overcome this issue, the VDIP \cite{vdip} was proposed that employs a VB-based approach in DIP. The posterior distribution $P$ is approximated using a trivial distribution $Q$ (e.g., Gaussian), and the distance measure is the KL divergence. The problem in Eq. (\ref{eq:dipe}) can be converted into the problem of minimizing the KL divergence
\vspace{-0.1cm}
\begin{equation}\label{DKL}
\begin{small}
\begin{aligned}
& D_{K L}\left(Q\left(k, u,  \xi_x, \xi_y\right) \| P\left(k, u,  \xi_x, \xi_y \mid s\right)\right) \\
= & \ln P\left(s\right)- \int Q\left(k\right) \ln \frac{P\left(k\right)}{Q\left(k\right)} dk
+\int Q\left( u\right) \ln Q\left(u\right) d u \\
& -\int Q\left(u\right) Q\left(\xi_x, \xi_y\right) \ln P\left(u \mid \xi_x, \xi_y\right) d u d \xi_x d \xi_y \\
& -\int Q\left(\xi_x, \xi_y\right) \ln \frac{P\left(\xi_x, \xi_y\right)}{Q\left(\xi_x, \xi_y\right)} d \xi_x d \xi_y \\
& -E_{Q\left( u\right)}\left[\ln P\left(s \mid u,k\right)\right],\\
\end{aligned}\end{small}
\end{equation}
where $D_{KL}$ represents the KL divergence, 
$L\left(k, u, \xi_x, \xi_y, s\right)$ is the variational lower bound, $P(k)$ is
set as the standard Gaussian distribution, and $\xi_x$ and $\xi_y$ are the variational parameters introduced for the convenience of calculation. The third line in Eq. (\ref{DKL}) is easy to calculate because $\ln P(s)$ is constant, and other terms only relate to the Gaussian distribution. Gaussian distributions can simplify derivation and implementation due to the continuity of their derivatives. The last line can be calculated by Monte Carlo estimation \cite{mc}. However, the fourth and fifth lines are hard because we don't know the distribution of $P\left(u \mid \xi_x, \xi_y\right)$ and $P\left(\xi_x, \xi_y\right)$. Thus $\xi_x$ and $\xi_y$ are introduced as variational parameters such that $P\left(u \mid \xi_x, \xi_y\right)$ is Gaussian, and $\xi_x$ and $\xi_y$ are related to image $u$. Then the fourth line in Eq. (\ref{DKL}) is easy to calculate because all terms are related to Gaussian distribution. The fifth line in Eq. (\ref{DKL}) is also addressed because the integral operation whose variables involve only $\xi_x$ and $\xi_y$ can be ignored \cite{vdip}.

How the distribution of $P\left(u \mid \xi_x, \xi_y\right)$ becomes Gaussian is explained as follows. The traditional image prior can be expressed as a super-Gaussian distribution, which can be expressed by the formula
\vspace{-0.2cm}
\begin{equation}\label{eq:p_gau}
\begin{small}
    P(u) = W \mathrm{exp}\left (-\frac{\rho(F_x(u))+\rho(F_y(u))}{2}\right ),
    \end{small}
\end{equation}
    where $W$ is the normalization coeffcient, and $\rho(\cdot)$ is the penalty function to constrain sparsity. $F_x(\cdot)$ and $F_y(\cdot)$ are gradient kernels $[-1,1]^T$ and $[-1,1]$. According to Babancan \textit{et al.} \cite{babaca}, the upper bound of $\rho(F_x(u))$ and of $\rho(F_y(u))$ are represented as
    \vspace{-0.1cm}
\begin{equation}\label{eq:inequation}
\begin{small}
\begin{aligned}
   \rho(F_x(u)) \leq \frac{1}{2}\xi_x(F_x(u))^2-\rho^*(\frac{1}{2}\xi_x),\\
    \rho(F_y(u)) \leq \frac{1}{2}\xi_y(F_y(u))^2-\rho^*(\frac{1}{2}\xi_y),
\end{aligned}
\end{small}
\end{equation}
where $\rho^*(\frac{1}{2}\xi_x)$ and $\rho^*(\frac{1}{2}\xi_y)$ denote the concave conjugates of $\rho(\sqrt{F_x(u)})$ and $\rho(\sqrt{F_y(u)})$, respectively, and $\xi_x$ and $\xi_y$ are the variational parameters.  Since $\frac{1}{2}\xi_x(F_x(u))^2-\rho^*(\frac{1}{2}\xi_x)$ and $\frac{1}{2}\xi_y(F_y(u))^2-\rho^*(\frac{1}{2}\xi_y)$ are both convex quadratic function, both of them has global minimum. The equation in Eq. (\ref{eq:inequation}) holds true when
\begin{equation}   
\begin{small}
\xi_x=\frac{\rho^{'}(F_x(u))}{\left| F_x(u)\right|},
\xi_y=\frac{\rho^{'}(F_y(u))}{\left| F_y(u)\right|},
\end{small}
\end{equation}
where $\rho^{'}(\cdot)$ is the dervation of $\rho(\cdot)$. If $u$ is determined, $\xi_x$ and $\xi_y$ can also be determined by $u$.

By replacing the $\rho(F_x(u))$ and $\rho(F_y(u))$ in Eq. (\ref{eq:p_gau}) with the corresponding right half of Eq. (\ref{eq:inequation}), $P(u|\xi_x,\xi_y)$ becomes a trivial Gaussian distribution 
\vspace{-0.1cm}
\begin{equation}\label{eq:Puxi_xxi_y)}
\begin{small}
\begin{aligned}
    P(u|\xi_x,\xi_y)=&W \mathrm{exp} \left (-\frac{\xi_x(F_x(u))^2+\xi_y(F_y(u))^2}{4}\right )\\
    &\cdot \mathrm{exp} \left (-\frac{\rho^*(\frac{1}{2}\xi_y)+\rho^*(\frac{1}{2}\xi_y)}{2}\right ).
    \end{aligned}
    \end{small}
\end{equation}
    Then according to Eq. (\ref{eq:Puxi_xxi_y)}), the fourth row of formula Eq. (\ref{DKL}) is easy to calculate.  Finally, aided by $\xi_x$ and $\xi_y$, the computational difficulties of Eq. (\ref{DKL}) are solved.

According to Eq. (\ref{eq:Puxi_xxi_y)}) and (\ref{DKL}), the variational
lower bound $\mathcal{L}\left(k, u, \xi_x, \xi_y, s\right)$ can be rewritten as
\begin{equation}
\begin{small}
\begin{aligned}
& \mathcal{L}\left(k, u, \xi_x, \xi_y, s\right) \\
= &\frac{1}{2}\sum_{i=1}^I \sum_{j=1}^J(2\ln S(u(m,n)-E^2(k(i,j))-S^2(k(i,j)))\\
& +\frac{1}{2} \sum_{m=1}^k \sum_{n=1}^N 2 \ln S \left( u(m, n)\right) \\
& -\frac{1}{4} \sum_{m=1}^k \sum_{n=1}^N E\left(\left(F_x\left( u\right)(m, n)\right)^2\right) E\left(\xi_x(m, n)\right) \\
& -\frac{1}{4} \sum_{m=1}^k \sum_{n=1}^N E\left(\left(F_y\left( u\right)(m, n)\right)^2\right) E\left(\xi_y(m, n)\right) \\
& +E_{Q\left( u\right)}\left[\ln P\left(s \mid  u, k\right)\right],\\
\end{aligned}
\end{small}
\end{equation}
\vspace{0.1cm}
where $(i, j)$ denotes the pixel index of kernel $k$, $S(\cdot)$ and $E(\cdot)$ are the standard deviation and the expectation of distribution $Q(\cdot)$, respectively, $(m, n)$ is the pixel index of $u$ and variational parameters $\xi_x$  and $\xi_y$.    

Given that $\ln P(s)$ remains constant and  $D_{KL}$
  holds a non-negative value in Eq. (\ref{DKL}), the task of minimizing  $D_{KL}$
  is reduced to minimizing the $-\mathcal{L}\left(k, u, \xi_x, \xi_y, s\right)$. In order to better minimize $-\mathcal{L}\left(k, u, \xi_x, \xi_y, s\right)$, VDIP used a fully-connected network as the kernel generator $G_k$ to get the distribution of kernels (like $S(k)$) and an encoder-decoder as the image generator $G_u$ to get the distribution of clean images (like $S(u)$ and $E(u)$). The blind deblurring of VDIP can be represented by the following formula
  \begin{equation} \label{vdipsol}
  \begin{small}
	\begin{cases}
 \theta_u, \theta_k=\mathop{\arg\min}\limits_{\theta_u, \theta_k} -\mathcal{L}\left(k, u, \xi_x, \xi_y, s\right), \\
        u=T^u_{\theta_u}(z_u), 
        k=T^k_{\theta_k}(z_k),	
	\end{cases}
  \end{small}
  \end{equation}
  \vspace{0.1cm}
 where  $-\mathcal{L}\left(k, u, \xi_x, \xi_y, s\right)$ is the loss function, $z_u$ and $z_k$ are  random noises fed into the image generator $T^u$ and the kernel generator $T^k$, respectively, and $\theta_u$ and $\theta_k$ are parameters of $T^u$ and $T^k$, respectively. The network architectures of $T^u$ and $T^k$ are the same as that of DIP in Eq. (\ref{dip_blind}). A frame diagram of VDIP for image blind deconvolution can be seen in Fig. \ref{figoverview}. Since VDIP imposes 
constraints on the generated images and kernels by taking into account the variance of the generated image, VDIP has a better deblurring effect than DIP.  

\vspace{-2mm}

\subsection{TGV Regularization}\label{sec3.2}
TGV regularization generalizes TV regularization \cite{25} by addressing the step effect issue of TV \cite{25}. It was demonstrated that TGV regularization can retain more texture information and generate more visually pleasing results than TV regularization. The TGV of the order $k$ is defined as
\begin{equation}
\begin{small}
\begin{aligned}
	{{\rm{TGV}}_{\gamma}^{k}(u)}=&\text{sup}\{\int _ \Omega u{\rm{div}}^{k}vdx\mid v\in C_{c}^{k}(\Omega,{\rm{Sym}}^{k}(\mathbb{R}^{d})),\\		
 &\Vert {\rm div}^{l}v \Vert_{\infty} \leq \gamma_{l},l=0,\cdots ,k-1\},
\end{aligned}
\end{small}
\end{equation}
where ${\rm{Sym}}^{k}(\mathbb{R}^{d})$ denotes the space of symmetric tensors of the order $k$ with arguments in $\mathbb{R}^{d}$, and $\gamma_{l}, l=0,\cdots ,k-1$ are fixed positive parameters. TGV can be interpreted as a “sparse”
penalization of optimal balancing from the first up to the $k-$th distributional derivative. 
Specially, when $k=1$, $\gamma=1$, $\rm TGV_{1}^{1}$ is equivalent to $\rm TV$ and when $k=2$ ,  $\gamma < 0$, 
\begin{equation}
\begin{small}
\begin{aligned}	
{{\rm{TGV}}_{\gamma}^{2}(u)}=&\text{sup}\{\int _ \Omega u{\rm {div}^{2}}\omega dx\mid \omega\in C_{c}^{2}(\Omega,S^{d*d}), \\
	&\Vert \omega \Vert_{\infty}\le \gamma_{0},\Vert {\rm div}\omega\Vert_{\infty}\le \gamma_{l},l=0,1,\cdots ,k-1\},
\end{aligned}
\end{small}
\end{equation}
which is called the second-order $\rm{TGV}$. In the case $k>2$, ${\rm TGV}_{\gamma}^{k}(u)$ is called the high-order $\rm{TGV}$. We have adopted the second-order TGV in our model due to the higher computational complexity involved with high-order $\rm{TGV}$, which involves the high-order divergence operators.

Since the $k$-th order TGV can be interpreted as a sparse penalization of optimal balancing from the first derivative to the $k$-th order derivative, the second-order TGV must contain the second-order derivative. Given an image $u$, $\operatorname{TGV}_{\gamma}^{2}(u)$ performs well in
penalization of $\nabla^2u$ in homogenous regions because
$\Vert \nabla^2u \Vert_1$ is considerably smaller than $\Vert \nabla u \Vert_1$, thus the undesirable noise and artifacts could be suppressed correspondingly in homogenous regions. In contrast, $\Vert \nabla^2u \Vert_1$ is
significantly larger than $\Vert \nabla u \Vert_1$ in the neighborhood of
edges. In this case, $\operatorname{TGV}_{\gamma}^{2}(u)$ has the capacity to suppress staircase-like artifacts and preserve
sharp edges.

The digital representation of images is mostly discrete, so in order to better apply TGV regularization to image processing, discrete TGV is introduced. According to 
\cite{tgvlisan, yinwotaotgv}, when $\operatorname{div} w=v$ in $(2)$, $U=  C_{c}^{2}(\Omega, \mathbb{R})$, $V= C_{c}^{2}\left(\Omega, \mathbb{R}^{2}\right)$ and $W= C_{c}^{2}\left(\Omega, \rm S^{2 \times 2}\right)$, the discretized ${\mathrm{TGV}}_{\gamma}^{2}$ can be written as
\begin{equation}
\begin{small}
\begin{aligned}
\operatorname{TGV}_{\gamma}^{2}(u)=&\max _{v \in V, w \in W}\{\langle u, \operatorname{div} v\rangle \mid \operatorname{div} w=v,\|w\|_{\infty} \leq \gamma_{0},\\
&\|v\|_{\infty} \leq \gamma_{1}\},
\end{aligned}
\end{small}
\end{equation}
where
$\operatorname{div} w=\left[\begin{array}{l}
	\partial_{x} w_{11}+\partial_{y} w_{12} \\
	\partial_{x} w_{21}+\partial_{y} w_{22}
\end{array}\right].$
The indicator function of a closed set $B$ can be defined as
$
\mathcal{I}_{B}= \begin{cases}0, & x \in B \\ \infty, & \text { else }\end{cases}.
$
Then according to the fact that $\mathcal{I}_{\{0\}}(\cdot)=-\min _{y}\langle y, \cdot\rangle$, the representation of the discrete $\mathrm{TGV}^{2}$ is as follows:
\vspace{-0.1cm}
\begin{equation}
\begin{small}
	\begin{aligned} 
		\operatorname{TGV}_{\gamma}^{2}(u) =&\min _{q \in V } \max _{\substack{\|w\|_{\infty} \leq \gamma_{0}, w \in W \\ \|v\|_{\infty} \leq \gamma_{1}, v \in V}} \langle u, \operatorname{div} v\rangle+\langle q, v-\operatorname{div} w\rangle \\ =&\min _{q \in V } \max _{\substack{\|w\|_{\infty} \leq \gamma_{0}, w \in W \\ \|v\|_{\infty} \leq \gamma_{1}, v \in V}}\langle-\nabla u, v\rangle+\langle q, v\rangle+\langle\overline{\mathcal{B}}(q), w\rangle. 
	\end{aligned}
  \end{small}
\end{equation}
Using the symmetry
property of the constraint $\{\|v\|_{\infty} \leq \gamma_{1}\}$ about zero and replacing $v$ with $-v$, the discrete form of $\mathrm{TGV}^{2}$ finally becomes
\vspace{-0.1cm}
\begin{equation}\label{eq:tgv2}
   \begin{small}
    \begin{aligned} \operatorname{TGV}_{\gamma}^{2}(u) =&\min _{q \in V } \max _{\substack{\|w\|_{\infty} \leq \gamma_{0}, w \in W \\ \|v\|_{\infty} \leq \gamma_{1}, v \in V}}\langle \nabla u-q, v \rangle+\langle\overline{\mathcal{B}}(q), w\rangle \\ =&\min _{q \in V} \gamma_{1}\|\nabla u-q\|_{1}+\gamma_{0}\|\overline{\mathcal{B}}(q)\|_{1},
    \end{aligned}  
        \end{small}
\end{equation}
where the operators $\nabla$ and $\overline{\mathcal{B}}$ can be represented by
\begin{equation}
    \begin{small}
	\begin{cases}
		&\nabla u=\left[\begin{array}{c}
			\nabla_{1} u \\
			\nabla_{2} u
		\end{array}\right] = \left[\begin{array}{c}
			\partial_{x} u \\
			\partial_{y} u
		\end{array}\right]\\       	 
		&\overline{\mathcal{B}}(q)=\left[\begin{array}{cc}
			\partial_{x} q_{1} 
			& \frac{1}{2}\left(\partial_{y} q_{1}+\partial_{x} q_{2}\right) \\
			\frac{1}{2}\left(\partial_{y} q_{1}+\partial_{x} q_{2}\right) 
			& \partial_{y} q_{2}
		\end{array}\right].
\end{cases}
\label{eq:derivation}
\end{small}
\end{equation}

\subsection{ADMM Algorithm}\label{sec3.3}
The ADMM algorithm \cite{admm1, admm2} is a classic approach for addressing optimization problems that have linear constraints. When faced with complex objective functions, it is challenging to find an optimal solution directly. However, by utilizing the ADMM algorithm, one can decompose a difficult optimization problem into a set of simpler sub-problems that can be iteratively solved to obtain the optimal solution. This process enables the ADMM algorithm to tackle challenging optimization problems. Typically, the optimization problems that ADMM algorithm can solve are
\begin{equation}
\begin{small}
	\begin{array}{l}
		 \min\limits_{y,z} f_{1}(y)+f_{2}(z) \\		 
		 \text{s.t. } y+Az+B=0,
	\end{array}
  \end{small}
\end{equation}
where $y \in \mathbb{R}^m$, $z \in \mathbb{R}^q $, $A \in \mathbb{R}^{m \times q}$, $B \in \mathbb{R}^m$ , $f_{1}: \mathbb{R}^p \rightarrow  \mathbb{R}$, $f_{2}: \mathbb{R}^q \rightarrow  \mathbb{R}$.
According to linear constraints $y+Az+B=0$, the Lagrange functional equation of the problem can be written as follows
\vspace{-0.1cm}
\begin{equation}
\begin{small}
\begin{aligned}
	\mathcal{L}_\xi(y, z, \lambda) 
	=&f_{1}(y)+f_{2}(z)+\frac{\xi}{2}\|y+Az+B\|_2^2\\
	&+\lambda^{\top}(y+Az+B),
\end{aligned}
\end{small}
\end{equation}
where $\lambda \in \mathbb{R}^{k}$, $\lambda$ is  referred to as the Lagrange multiplier and $\xi$ represents the penalty parameter. 
Then, the variables $y$,$z$ and $\psi$ are iteratively updated according to the following equation
\vspace{-0.1cm}
\begin{equation}
\begin{small}
\begin{aligned}
&y^{(k+1)}=\mathop{\arg\min}\limits_{y} f_{1}(y)+\frac{\xi}{2}\Vert y+Az^{k}+B+\psi^{k}\Vert_{2}^{2}, \\
	&z^{(k+1)}=\mathop{\arg\min}\limits_{z} f_{2}(z)+\frac{\xi}{2}\Vert y^{k+1}+Az+B+\psi^{k}\Vert_{2}^{2},  \\
	&\psi^{(k+1)}=\psi^{(k)}+y^{(k+1)}+Az^{(k+1)}+B,
\end{aligned}
\end{small}
\end{equation}
where $\psi=\frac{\xi}{\lambda}$. Lastly, the best solution for $y$ and $z$ can be obtained by continuously updating $y$, $z$ and $\psi$ alternatively.

\section{Our Model}\label{sec4}

Here, we propose a single-image blind deconvolution model by combining VDIP and TGV. The loss function comes from the derivation of the VB. Our method is different from DIP-based methods. While several DIP-based methods have explored combining DIP with regularization, such as DIP-TV \cite{33} and DIP-VBTV \cite{34}, to the best of our knowledge, no one has so far attempted to combine the VDIP with any forms of regularization for single-image blind deconvolution. Moreover, adding the TGV regularization into VDIP preserves details and captures more information when the blur kernel is large and the image size is small.

The ADMM algorithm is used to solve the model, and it transforms the problem into solving 4 sub-problems, in which two networks $T^u$ (capturing the prior information of clean images) and $T^k$ (capturing the prior information to blur kernels) are trained to solve the second sub-problem. Two networks share a loss function and update iteratively at the same time. The optimization strategy we provide can be thought of as a major modification of  \cite{ourmethod}. \cite{ourmethod} utilizes a pre-trained GAN in conjunction with a prior $R(z)$, and solves the resulting optimization problem through the ADMM algorithm. One key difference between our approach and previous deep learning-based methods like \cite{ourmethod} is that we achieve high-quality results without the need for large datasets or extensive training. Traditional variational models have typically not incorporated neural networks in conjunction with the ADMM algorithm, whereas our approach leverages them to significantly enhance image quality. Specifically, we adapt the objective function of a subproblem in the ADMM algorithm as the loss function according to the VB method. The overview of our method is clearly displayed in Fig. \ref{figoverview}.

\subsection{VDIP-TGV}\label{sec4.1}
In this paper, we propose a general single-image blind deconvolution model called VDIP-TGV, which combines the penalty term ${\rm TGV}_{\gamma}^{2}(u)$ and the fidelity term ${\Vert k\otimes u-s \Vert}^{2}_{2}$. According to Eq. (\ref{eq:tgv2}) for the second-order TGV, the
objective function in Eq. (\ref{eq:dipe}) can be rewritten as
\vspace{-0.2cm}
 \begin{equation}
 \begin{small}
\mathop{\min}\limits_{u,k}\frac{\beta}{2}\Vert k\otimes u-s \Vert^{2}_{2}+ {\rm TGV}_{\gamma}^{2}(u),
 \label{ourproblem}
  \end{small}
\end{equation}
where ${\rm TGV}_{\gamma}^{2}(u)=\gamma_{1}\|\nabla u-q\|_{1}+\gamma_{0}\|\overline{\mathcal{B}}(q)\|_{1}$, and $\gamma_{1}$ and $\gamma_{0}$ are positive numbers.
In order to facilitate the calculation, we approximate directional
derivatives $\nabla_{1}^{u}$ and $\nabla_{2}^{u}$ in Eq. (\ref{eq:derivation}) by $D_{1}^{u}$ and $D_{2}^{u}$ like \cite{yinwotaotgv}, where $D_{1}^{u}$ and $D_{2}^{u}$ are  the forward finite difference operators along the $x$ axis and $y$ axis, respectively. After replacing $\nabla u$ with $Du$ and approximating $\overline{\mathcal{B}}(q)$ by $\mathcal{B}(q)$, Eq. (\ref{eq:derivation}) can be rewritten as 
\vspace{-0.1cm}
\begin{equation}
\begin{small}
	\begin{cases}	&Du=\left[\begin{array}{c}
			D_{1}u \\
			D_{2}u
	\end{array}\right],\\       	 
		&\mathcal{B}(q)=\left[\begin{array}{cc}
		D_{1}q_{1}
		& \frac{1}{2}\left(D_{2}q_{1}+D_{1}q_{2}\right) \\
		\frac{1}{2}\left(D_{2}q_{1}+D_{1}q_{2}\right) 
		& D_{2}q_{2}
	\end{array}\right].
\end{cases}
\end{small}
\end{equation}
Our approach uses a generator $T^{u}_{\theta_u}$ to estimate blur kernels and a generator $T^{k}_{\theta_k}$ to estimate clean images. The network structures of both generators are the same as those of VDIP. The procedure for solving our method can be written as
\vspace{-0.1cm}
 \begin{equation}
  \begin{small}
	\begin{cases}
		\theta_u, \theta_k=\mathop{\arg\min}\limits_{\theta_u,\theta_k}\frac{\beta}{2}\Vert k\otimes u-s \Vert^{2}_{2}+ {\rm TGV}_{\gamma}^{2}(u),       \\
		{\rm TGV}_{\gamma}^{2}(u)=\gamma_{1}\|Du-q\|_{1}+\gamma_{0}\|\mathcal{B}(q)\|_{1}, \gamma_{1}>0, \gamma_{0}>0,\\
		u=T^{u}_{\theta_u}(z_u),
  k=T^{k}_{\theta_k}(z_k),
	\end{cases}
 \label{tgvsolving}
 \end{small}
\end{equation}
where $z_u$ and $z_k$ are the inputs of $T^{u}$ with parameters $\theta_u$ and $T^{k}$ with parameters $\theta_k$, respectively. Treating the objective function of the first row in Eq. (\ref{tgvsolving}) as a loss function like DIP and VDIP does not work because it is difficult to calculate $q$ according to the output of each iteration ($u$).  Considering that the ADMM algorithm is famous for dividing and conquering the complex problem, we use the ADMM algorithm to solve the objective function.

\subsection{Optimizing VDIP-TGV}\label{sec4.2}

Algorithm \ref{algorithm} gives a brief summary of our optimization strategy. According to the ADMM algorithm, we introduce 
auxiliary variables:
\begin{equation}\nonumber
\begin{small}
\begin{aligned}
	&g=\left[\begin{array}{cc}
		g_{1} \\
		g_{2}
	\end{array}\right] \in V,
	&h=\left[\begin{array}{cc}
		h_{1} 
		&h_{3}\\
		h_{3}
		&h_{2}
	\end{array}\right] \in W,
\end{aligned}
\end{small}
\end{equation}
where $V= C_{c}^{2}\left(\Omega, \mathbb{R}^{2}\right)$ and $W= C_{c}^{2}\left(\Omega, \rm S^{2 \times 2}\right)$, and rewrite Eq. (\ref{ourproblem}) as
\begin{equation}
\begin{small}
	\begin{aligned}
		&\underset{g,h,u,q}{\text {min}} \frac{\beta}{2}\Vert k\otimes u-s \Vert_{2}^{2}+\gamma_{1}\|g\|_{1}+\gamma_{0}\|h\|_{1}\\
		&\text { s.t. }  g=Du-q, h=\mathcal{B}(q).
	\end{aligned}
 \label{Largran}
 \end{small}
\end{equation}Then after introducing the Lagrangian multipliers $\widetilde{g}$, $\widetilde{h}$ and  penalty parameters $\varphi_{1}$, $\varphi_{2}>0$, and denoting $\widetilde{g}=\frac{\varphi_{1}}{g}$,  $\widetilde{h}=\frac{\varphi_{2}}{h}$, the augmented Lagrangian equation for the aforementioned minimization issue (\ref{Largran}) is therefore obtained as follows:
\begin{equation}
\begin{small}
    \begin{aligned}
    &\underset{g,h,u,q}{\text {min}}\frac{\beta}{2}\Vert k\otimes u-s \Vert_{2}^{2}+\gamma_{1}\|g\|_{1}+\gamma_{0}\|h\|_{1}\\
    &+\frac{\varphi_{1}}{2}\Vert g-(Du-q)-\widetilde{g}\Vert_{2}^{2}
    +\frac{\varphi_{2}}{2}\Vert h-\mathcal{B}(q)-\widetilde{h}\Vert_{2}^{2}.
    \end{aligned}
    \end{small}
\end{equation}Finally, the following five subproblems are resolved alternatively to arrive at the optimal solution.

$\bullet$  $g$-subproblem
\begin{equation}
\begin{small}
g^{n+1}=\mathop{\arg\min}\limits_{g}\gamma_{1}\Vert g \Vert_{1}+\frac{\varphi_{1}}{2}\Vert g-(Du^{n}-q^{n})-\widetilde{g}^{n}\Vert_{2}^{2}.
\end{small}
\end{equation}

Since the $g$-subproblem is elementwise separable, the solution to the $g$-subproblem reads as
\begin{equation}\label{eq:gsubproblem}
\begin{small}
	g^{n+1}(l)=\gamma_{1}(\mathcal{S}\mathcal{H}_{2}\left(D u^{n}(l)-q^{n}(l)+\tilde{g}^{n}(l), \gamma_{1} / \varphi_{1}\right)), \quad l \in \Omega,
 \end{small}
\end{equation}
where $g^{n+1}(l)$ represents $g^{n+1}$ located at $l \in \Omega$, and the isotropic shrinkage operator $\mathcal{S}\mathcal{H}_{2}(a, \varphi)$ is defined as
\begin{equation}\nonumber
\begin{small}
\mathcal{S}\mathcal{H}_{2}(a, \varphi)= \begin{cases}0, & a=0, \\ \left(\|a\|_{2}-\varphi\right) \frac{a}{\|a\|_{2}}, & a \neq 0.\end{cases}
\end{small}
\end{equation}

$\bullet$  $h$-subproblem
\begin{equation}\label{eq:hsubproblem}
\begin{small}
h^{n+1}=\mathop{\arg\min}\limits_{h}\gamma_{0}\Vert h \Vert_{1}+\frac{\varphi_{2}}{2}\Vert h-(\mathcal{B}(q^{n}))-\widetilde{h}^{n}\Vert_{2}^{2}.
\end{small}
\end{equation}

Likewise, the solution to the $h$-subproblem is as follows
\begin{equation}
\begin{small}
	h^{n+1}(l)=\gamma_{0}(\mathcal{S}_{F}(\mathcal{B}\left(q^{n}\right)(l)+\tilde{h}^{n}(l), \gamma_{0} / \varphi_{2})), \quad l \in \Omega,
 \end{small}
\end{equation}
where $h^{n+1}(l)\in {\rm{S}}^{2*2}$ is the component of $h^{n+1}$corresponding to the pixel $l\in \Omega$ and 
\begin{equation}\nonumber
\begin{small}
\mathcal{S}_{F}(b, \varphi)= \begin{cases}0, & b=0, \\ \left(\|b\|_{F}-\varphi\right) \frac{b}{\|b\|_{F}}, & b \neq 0.\end{cases}
\end{small}
\end{equation}

$\bullet$  $u$-subproblem
\begin{equation}\label{eq:u}
\begin{small}
u^{n+1}=\mathop{\arg\min}\limits_{u}\frac{\beta}{2}\Vert k\otimes u-s \Vert_{2}^{2}+\frac{\varphi_{1}}{2}\Vert g^{n+1}-(Du-q^{n})-\widetilde{g}^{n}\Vert_{2}^{2}.
\end{small}
\end{equation}

\begin{figure*}[htbp]
	\centering
	\includegraphics[scale=0.58]{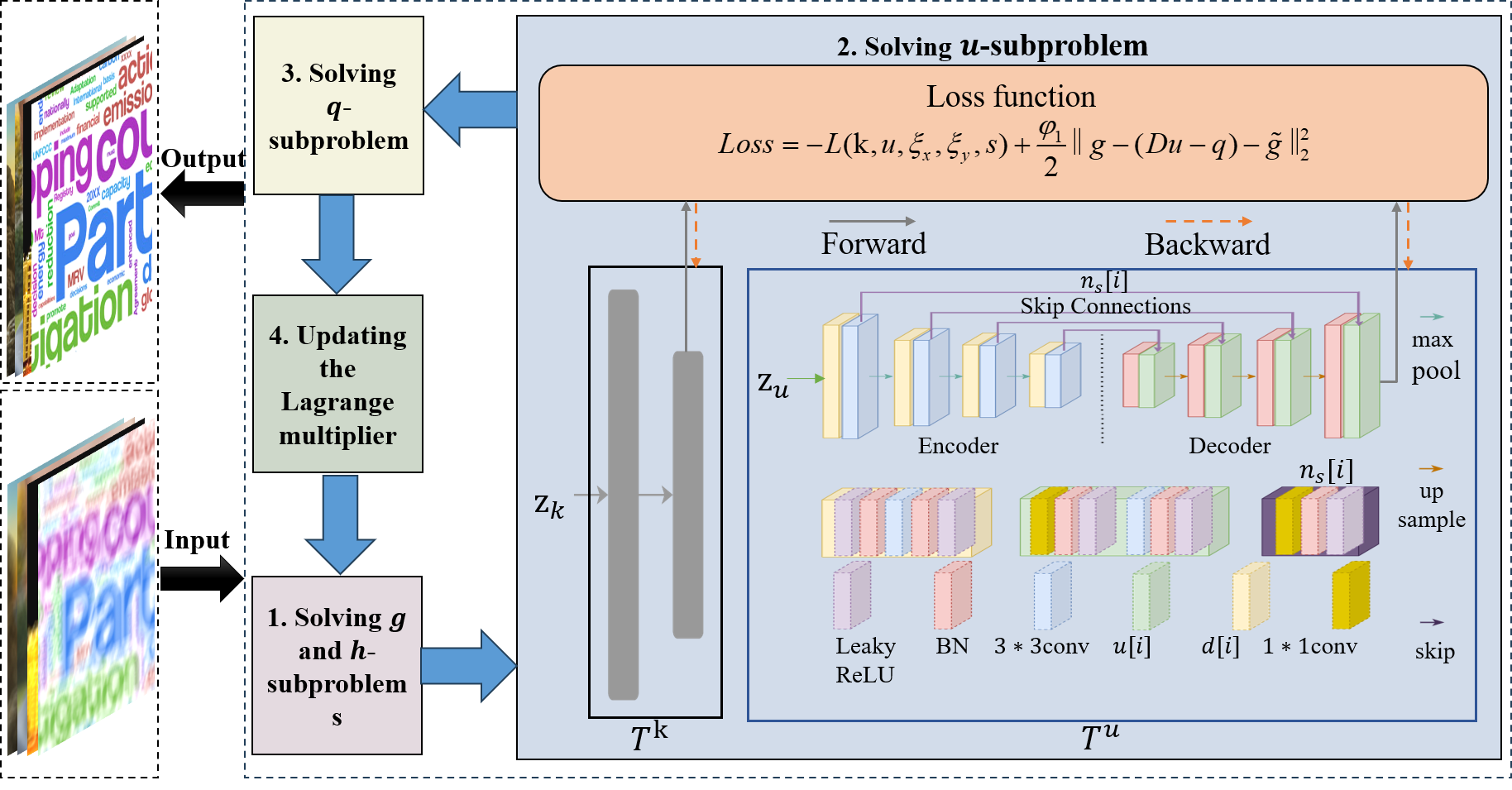}
	\caption{The flowchart of the proposed VDIP-TGV. The ADMM algorithm transforms the problem into solving 4 sub-problems, in which two networks $T^u$ (which captures the prior information for clean images) and $T^k$ (which captures the prior information for blur kernel) are used to solve the second sub-problem. The inputs to $T^u$ and $T^k$ are randomly initialized noise images, denoted by the symbols $z_u$ and $z_k$, respectively. The network architecture of $T^u$ is the same as that of the VDIP and DIP, using an autoencoder to output the deblurred image. $T^k$ uses a fully connected layer to output blur kernel, which is the same as VDIP. Both $T^u$ and $T^k$ use the same loss function. The determination of the loss function is based on the VB method, see Section \ref{sec4.2} for details.}
 \label{oursdetail}
\end{figure*}

In order to solve this objective function, we split it into two parts: $\frac{\beta}{2}\Vert k\otimes u-s \Vert_{2}^{2}$ and the rest. Drawing inspiration from the VDIP, we minimize the first part using the VB approach so that we could take advantage of the variance constraints of the images to avoid the issues of sparse MAP. VB involves utilizing a simple distribution, such as a Gaussian distribution, denoted as $Q(k, u, \xi_x, \xi_y)$, to approximate the posterior distribution $P(k, u, \xi_x, \xi_y\|s)$. This approximation is achieved by minimizing the KL divergence, as represented in Eq. (\ref{DKL}). Then minimizing the first part  in Eq. (\ref{eq:u}) is  equivalent to minimizing the term 
 $-\mathcal{L}\left(k, u, \xi_x, \xi_y, s\right)$
in Eq. (\ref{DKL}). Since solving the $u$ subproblem is equivalent to minimizing the sum of $-\mathcal{L}\left(k, u, \xi_x, \xi_y, s\right)$ and the second part, the sum of $-\mathcal{L}\left(k, u, \xi_x, \xi_y, s\right)$ and the second part is viewed as our loss function. After a few iterations, we can get the output of the network, which is the solution to the $u$-subproblem. The $u$-subproblem's resolution procedure is as follows:
 \begin{equation}
  \begin{small}
 \begin{cases}
 \begin{aligned}
		\theta_u, \theta_k=&\mathop{\arg\min}\limits_{\theta_u,\theta_k}-\mathcal{L}\left(k, u, \xi_x, \xi_y, s\right)\\
  &+\frac{\varphi_{1}}{2}\Vert g^{n+1}-(Du-q^{n})-\widetilde{g}^{n}\Vert_{2}^{2},     
       \end{aligned}
       \\
		u=T^{u}_{\theta_u}(z_u),
  k=T^{k}_{\theta_k}(z_k),
  \end{cases}
 \label{usolve}
  \end{small}
\end{equation}
from the corresponding subproblem. The network architectures of $T^u$ and $T^k$ are the same as those of VDIP in Eq. (\ref{vdipsol}), which can be seen in Fig. \ref{oursdetail}.

$\bullet$  $q_{1}$ and $q_{2}$-subproblems
\begin{equation}
\begin{small}
\begin{aligned}
q^{n+1}=&\mathop{\arg\min}\limits_{q}\frac{\varphi_{1}}{2}\Vert g^{n+1}-(Du^{n+1}-q)-\widetilde{g}^{n}\Vert_{2}^{2}\\
&+\frac{\varphi_{2}}{2}\Vert h^{n+1}- \mathcal{B}(q)-\widetilde{h}^{n}\Vert_{2}^{2}.
\end{aligned}
\end{small}
\end{equation}

After taking the derivative of the objective function of this subproblem and setting it to 0, we can get the following equation
\begin{equation}
\begin{small}
\left\{
\begin{aligned}
	0=&\gamma_{1}\varphi_{1}(q_{1}-D_{1}u^{n+1}+g_{1}^{n+1}-\widetilde{g}_{1}^{n})\\
	&+\gamma_{0}\varphi_{2}(D_{1}^{T}(D_{1}q_{1}-h_{1}^{n+1}
	+\widetilde{h}_{1}^{n})\\
	&+\frac{1}{2}D_{2}^{T}(D_{2}q_{1}+D_{1}q_{2}-2h_{3}^{n+1}+2\widetilde{h}_{3}^{n})),\\
	0=&\gamma_{1}\varphi_{1}(q_{2}-D_{2}u^{n+1}+g_{2}^{n+1}-\widetilde{g}_{2}^{n})\\
	&+\gamma_{0}\varphi_{2}(D_{2}^{T}(D_{2}q_{2}-h_{2}^{n+1}+\widetilde{h}_{2}^{n})\\
	&+\frac{1}{2}D_{1}^{T}(D_{1}q_{2}+D_{2}q_{1}-2h_{3}^{n+1}+2\widetilde{h}_{3}^{n})).
\end{aligned}
\right.
\end{small}
\end{equation}
After the transformation  and basic arithmetic, we have
\begin{equation}\label{eq:qsubproblem}
\begin{small}
	\left\{
	\begin{aligned}
		q_{1}=&\frac{k_{1}}{k_{2}}, q_{2}=\frac{k_{3}}{k_{4}},\\
		k_{1}=&\gamma_{1}\varphi_{1}( D_{1}u^{n+1}-g_{1}^{n+1}+\widetilde{g}_{1}^{n} )+\gamma_{0}\varphi_{2} (D_{1}^{T}(h_{1}^{n+1}-\widetilde{h}_{1}^{n})\\
		+&D_{2}^{T}(h_{3}^{n+1}-\widetilde{h}_{3}^{n}-\frac{1}{2}D_{1}q_{2}^{n} )),\\
		k_{2}=&{\gamma_{1}\varphi_{1}+\gamma_{0}\varphi_{2}(D_{1}^{T}D_{1}+\frac{1}{2}D_{2}^{T}D_{2})},\\
		k_{3}=&\gamma_{1}\varphi_{1}( D_{2}u^{n+1}-g_{2}^{n+1}+\widetilde{g}_{2}^{n} )+\gamma_{0}\varphi_{2} (D_{2}^{T}(h_{2}^{n+1}-\widetilde{h}_{2}^{n})\\
		+&D_{1}^{T}(h_{3}^{n+1}-\widetilde{h}_{3}^{n}-\frac{1}{2}D_{2}q_{1}^{n} )),\\
		k_{4}=&{\gamma_{1}\varphi_{1}+\gamma_{0}\varphi_{2}(D_{2}^{T}D_{2}+\frac{1}{2}D_{1}^{T}D_{1})}.
	\end{aligned}
	\right.
 \end{small}
\end{equation}

$\bullet$  $\widetilde{g}$ and  $\widetilde{h}$-subproblems

The Lagrange multipliers are updated as follows
\begin{equation}\label{eq:Lsubproblem}
\begin{small}
	\left\{
	\begin{aligned}
		\widetilde{g}^{n+1}&=\widetilde{g}^{n}+\mu(Du^{n+1}-q^{n+1}-g^{n+1}),\\
		\widetilde{h}^{n+1}&=\widetilde{h}^{n}+\mu(\mathcal{B}(q^{n+1})-h^{n+1})	.
	\end{aligned}
	\right.
 \end{small}
\end{equation}
By iteratively updating $g$, $h$, $u$, $p$ and Lagrange multipliers  $\widetilde{g}$ and $\widetilde{h}$, the optimal solution $u$ is obtained. 
\begin{algorithm}	
\renewcommand{\algorithmicrequire}{\textbf{Input:}}	\renewcommand{\algorithmicensure}{\textbf{Output:}}
	\caption{}
	\label{alg:1}
	\begin{algorithmic}[1]
		\STATE  Choose $\gamma_{0}$, $\gamma_{1}$, $\varphi_{1}$, $\varphi_{2}$, $\beta$, $\mu$;
		
		Initialize $u_{0}$, $q_{1}^{0}$, $q_{2}^{0}$, $g_{j}^{0}$; $\widetilde{g}_{j}^{0}$, $(j=1,2)$, $h_{j}^{0}$, $\widetilde{h}_{j}^{0}, (j=1,2,3)$; 	
		
		Set $t$ as the maximum number of iterations.	
		\FORALL{$n=0,1,2,3,...,t$}
		\STATE Update $g^{n}$ based on Eq. (\ref{eq:gsubproblem});
		\STATE Update $h^{n}$ based on Eq. (\ref{eq:hsubproblem});
            \STATE Update  $k^{n}$ and $u^{n}$ based on Eq. (\ref{usolve});
		\STATE Update $q_{1}^{n}$, $q_{2}^{n}$ based on Eq. (\ref{eq:qsubproblem});
		\STATE Update $\widetilde{g}^{n}$, $\widetilde{h}^{n}$ based on Eq. (\ref{eq:Lsubproblem});
		\STATE $n=n+1$.
		\ENDFOR
		\STATE $U=u^{n}$.
		\RETURN $U$.
	\end{algorithmic} 
 \label{algorithm}
\end{algorithm}

\section{Experiment}\label{sec5}
In this section, we provide systematic quantitative and qualitative experiments to showcase the efficacy of our approach relative to other state-of-the-art methods.

\subsection{Implementation Details}\label{sec5.1}
\begin{figure*}[htb]
				\begin{minipage}[t]{0.16\linewidth}
			\centering
			\vspace{1pt}
			\centerline{\includegraphics[width=2.95cm,height=2.95cm]{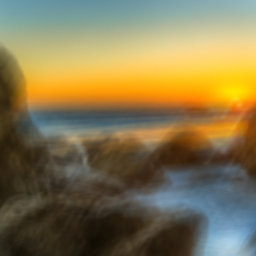}}
			\vspace{1pt}
			\centerline{\includegraphics[width=2.95cm,height=2.95cm]{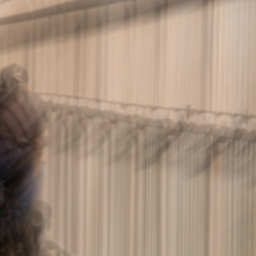}}
			\vspace{1pt}
			\centerline{\includegraphics[width=2.95cm,height=2.95cm]{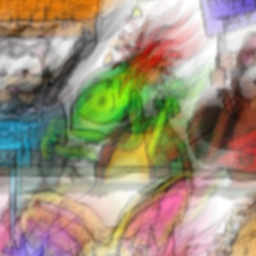}}
			\vspace{1pt}
			\centerline{\includegraphics[width=2.95cm,height=2.95cm]{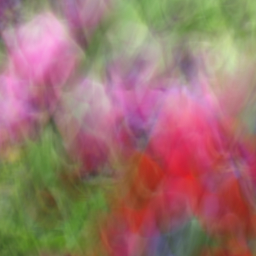}}
			\vspace{-3pt}
			\centerline{\small Blur}
		\end{minipage}
			\begin{minipage}[t]{0.16\linewidth}
			\centering
			\vspace{1pt}
			\centerline{\includegraphics[width=2.95cm,height=2.95cm]{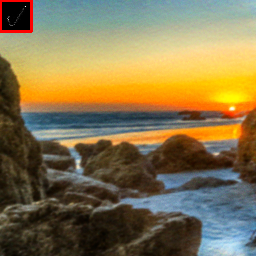}}
			\vspace{1pt}
			\centerline{\includegraphics[width=2.95cm,height=2.95cm]{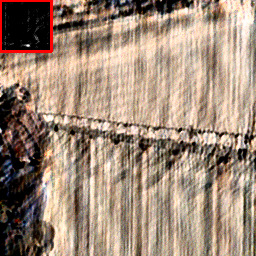}}
			\vspace{1pt}
			\centerline{\includegraphics[width=2.95cm,height=2.95cm]{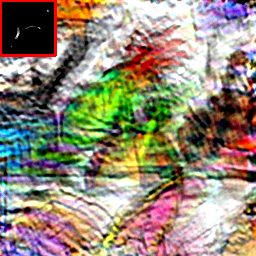}}
			\vspace{1pt}
			\centerline{\includegraphics[width=2.95cm,height=2.95cm]{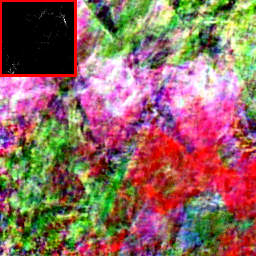}}
			\vspace{-3pt}
			\centerline{\small Ren \textit{et al.} DIP \cite{dip37}}
		\end{minipage}
			\begin{minipage}[t]{0.16\linewidth}
				\centering
				\vspace{1pt}
				\centerline{\includegraphics[width=2.95cm,height=2.95cm]{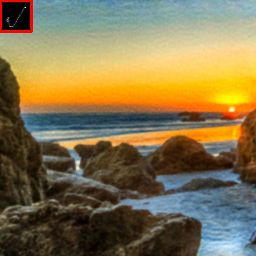}}
					\vspace{1pt}
				\centerline{\includegraphics[width=2.95cm,height=2.95cm]{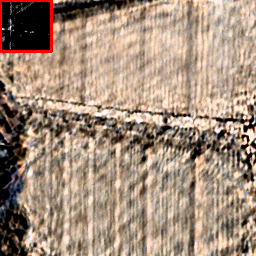}}
				\vspace{1pt}
				\centerline{\includegraphics[width=2.95cm,height=2.95cm]{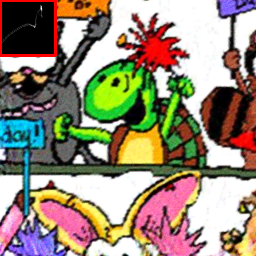}}
					\vspace{1pt}
				\centerline{\includegraphics[width=2.95cm,height=2.95cm]{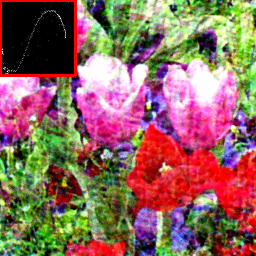}}
				\vspace{-3pt}
				\centerline{\small Dong \textit{et al.} VDIP\cite{vdip}}
			\end{minipage}
				\begin{minipage}[t]{0.16\linewidth}
				\centering
				\vspace{1pt}
				\centerline{\includegraphics[width=2.95cm,height=2.95cm]{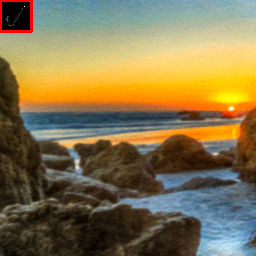}}
				\vspace{1pt}
				\centerline{\includegraphics[width=2.95cm,height=2.95cm]{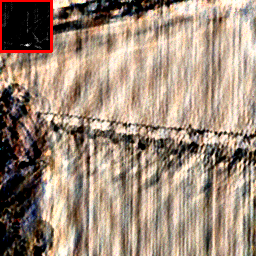}}
				\vspace{1pt}
				\centerline{\includegraphics[width=2.95cm,height=2.95cm]{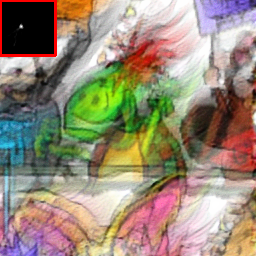}}
				\vspace{1pt}
				\centerline{\includegraphics[width=2.95cm,height=2.95cm]{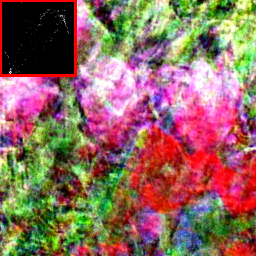}}
				\vspace{-3pt}
				\centerline{\small DIP-TGV}
			\end{minipage}			
			\begin{minipage}[t]{0.16\linewidth}
			\centering
			\vspace{1pt}
			\centerline{\includegraphics[width=2.95cm,height=2.95cm]{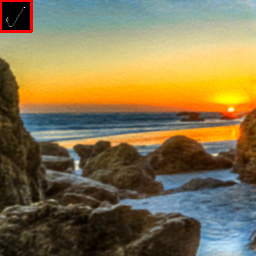}}
			\vspace{1pt}
			\centerline{\includegraphics[width=2.95cm,height=2.95cm]{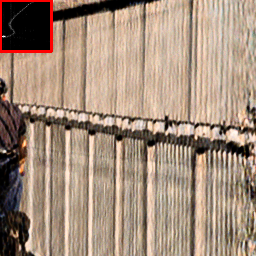}}
			\vspace{1pt}
			\centerline{\includegraphics[width=2.95cm,height=2.95cm]{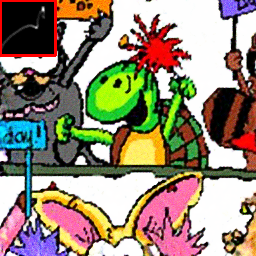}}
			\vspace{1pt}
			\centerline{\includegraphics[width=2.95cm,height=2.95cm]{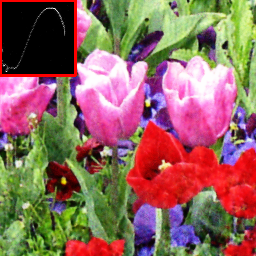}}
			\vspace{-3pt}
			\centerline{\small OURS (VDIP-TGV)}
		\end{minipage}
			\begin{minipage}[t]{0.16\linewidth}
			\centering
			\vspace{1pt}
			\centerline{\includegraphics[width=2.95cm,height=2.95cm]{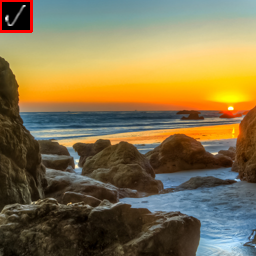}}
			\vspace{1pt}
			\centerline{\includegraphics[width=2.95cm,height=2.95cm]{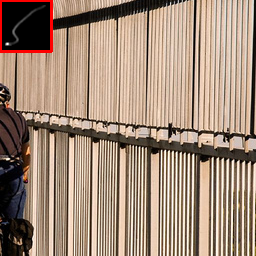}}
			\vspace{1pt}
			\centerline{\includegraphics[width=2.95cm,height=2.95cm]{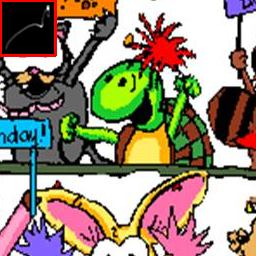}}
			\vspace{1pt}
			\centerline{\includegraphics[width=2.95cm,height=2.95cm]{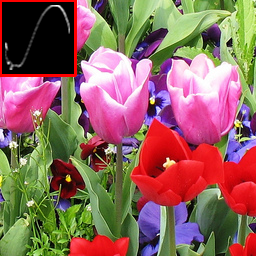}}
			\vspace{-3pt}
			\centerline{\small Clean}
		\end{minipage}		
		\caption{Qualitative comparisons between DIP-TGV and VDIP-TGV on four images.}
  \label{img: ablation experiment}
  \vspace{-0.4cm}
	\end{figure*}

\textbf{Datasets.} The experiment was conducted on two datasets, the dataset collected by Lai \textit{et al.} \cite{lai} and the dataset collected by Kohler \textit{et al.} \cite{ko}. The dataset collected by Lai \textit{et al.} \cite{lai} includes 200 images, 100 of which are synthetic images and the rest are real images. The 100 synthetic images are convolved by four different sizes of blur kernels. These images belong to different categories such as Manmade, Natural, People, Saturated, and Text.
The dataset collected by Kohler \textit{et al.} \cite{ko} is made by randomly selecting 12 blurry trajectories that imitate the camera shake, and applying them to 4 images, resulting in 48 blurry images.

\textbf{Methods.} We contrast our proposed VDIP-TGV with several other methods, including conventional methods such as 
Krishnan \textit{et al.} \cite{vdip41}, Pan \textit{et al.} \cite{vdip12}, Wen \textit{et al.} \cite{vdip67}, deep learning-based methods such as
Kupyn \textit{et al.} \cite{vdip30}, Zamir \textit{et al.} \cite{vdip69} and Chen \textit{et al.} \cite{vdip70}, and DIP-based methods such as DIP \cite{dip37} and VDIP \cite{vdip}. All these works were published in flagship journals in the field of image processing.
	
\textbf{Evaluation metrics.} We use both reference and no-reference image quantitative assessment metrics to evaluate our algorithm.  For reference metrics, we use Peak Signal-to-Noise Ratio (PSNR), Structural Similarity Index Measurement (SSIM) \cite{psnr39}, and Mean Squared Error (MSE) \cite{error}. When the PSNR value is larger or the SSIM value is closer to 1, the image quality after deblurring is considered to be higher. The smaller the error is, the more accurate the estimation of blur kernel is. For no-reference assessment metrics, we use Naturalness Image Quality Evaluator (NIQE) \cite{niqe}, Blind/Referenceless
Image Spatial Quality Evaluator (BRISQUE) \cite{brisque}, and Perception based Image Quality EvaluatorPIQE (PIQE) \cite{piqe}. The smaller the three indicators, the better quality the images have.

\textbf{Setting.} Our method uses networks $T^u$ and $T^k$ to capture the prior information for clean images and blur kernels, respectively. For a fair comparison, the architectures of $T^u$ and $T^k$ used in VDIP-TGV are the same as those in VDIP, where $T^u$ uses the autoencoder to yield a deblurring image, $T^k$ uses the fully connected layer to estimate a blur kernel, and the structural details of $T^u$ and $T^k$ can be seen in Fig \ref{oursdetail}. Similar to DIP \cite{dip37} and VDIP \cite{vdip}, the total count of iterative steps is prescribed as 5,000 and the learning rate of the image generator and of the kernel generator are set as $1 \times 10^{-2}$ and $1 \times 10^{-4}$, respectively. All experiments are conducted using PyCharm 2021 on a PC equipped with an Intel Core i7-RTX 3050 on Windows 10.

\subsection{Analyzing the Effectiveness of VDIP-TGV}
The highlight of our method is the employment of VDIP and the introduction of TGV. To confirm the effectiveness of VDIP-TGV, we compare VDIP-TGV with DIP \cite{dip37}, VDIP \cite{vdip}, and DIP-TGV. We randomly select 4 blurry images from Lai \textit{et al.} \cite{lai} convoluted by kernels of different sizes for visualization and perform numerical experiments on them. The recovery images are shown in Fig. \ref{img: ablation experiment}, and the numerical results are included in Tab. \ref{tabel1}. For a unified presentation, we select test images by cropping $256\times256$ pixels from the middle of the original test images.

\begin{figure}[ht]
		\begin{minipage}[t]{0.19\linewidth}
			\centering
	\centerline{\includegraphics[width=1.75cm,height=1.75cm]{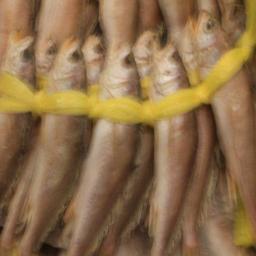}}
			\vspace{-8pt}
			\centerline{\tiny Real}
			\vspace{4.2pt}
			\centerline{\includegraphics[width=1.75cm,height=1.75cm]{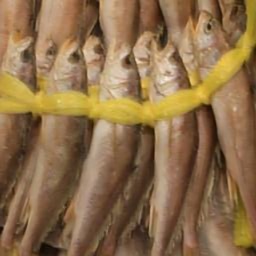}}
			\vspace{-8pt}
			\centerline{\tiny Zamir \textit{et al.} \cite{vdip69}}
		\end{minipage}
		\begin{minipage}[t]{0.19\linewidth}
			\centering
			\centerline{\includegraphics[width=1.75cm,height=1.75cm]{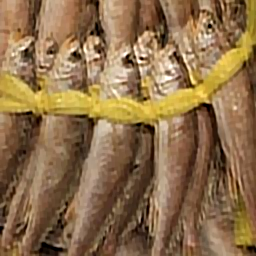}}
			\vspace{-8pt}
			\centerline{\tiny Krishnan \textit{et al.} \cite{vdip41}}
			\vspace{3.4pt}
			\centerline{\includegraphics[width=1.75cm,height=1.75cm]{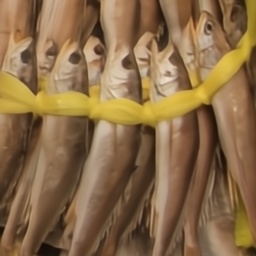}}
			\vspace{-8pt}
			\centerline{\tiny Chen \textit{et al.} \cite{vdip70}}
		\end{minipage}
		\begin{minipage}[t]{0.19\linewidth}
			\centering
			\centerline{\includegraphics[width=1.75cm,height=1.75cm]{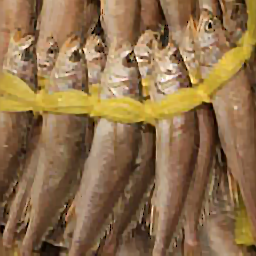}}
			\vspace{- 8pt}
			\centerline{\tiny Pan \textit{et al.} \cite{vdip12}}
			\vspace{3.5pt}
			\centerline{\includegraphics[width=1.75cm,height=1.75cm]{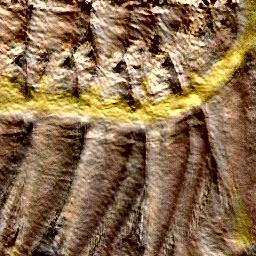}}
			\vspace{-8pt}
			\centerline{\tiny Ren \textit{et al.} DIP\cite{dip37}}
		\end{minipage}
		\begin{minipage}[t]{0.19\linewidth}
			\centering
			\centerline{\includegraphics[width=1.75cm,height=1.75cm]{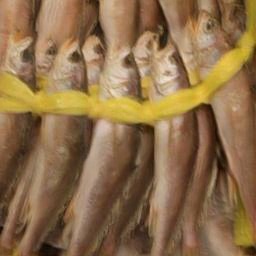}}
			\vspace{-8pt}
			\centerline{\tiny Kupyn \textit{et al.} \cite{vdip30}}
			\vspace{3.3pt}
			\centerline{\includegraphics[width=1.75cm,height=1.75cm]{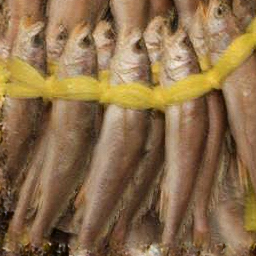}}
			\vspace{-8pt}
			\centerline{\tiny Dong \textit{et al.} VDIP\cite{vdip}}
		\end{minipage}
		\begin{minipage}[t]{0.19\linewidth}
			\centering
			\centerline{\includegraphics[width=1.75cm,height=1.75cm]{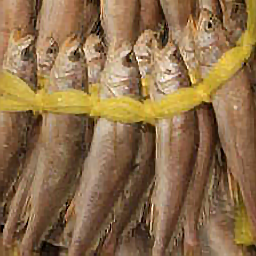}}
			\vspace{-8pt}
			\centerline{\tiny Wen \textit{et al.} \cite{vdip67}}
			\vspace{3.5pt}
			\centerline{\includegraphics[width=1.75cm,height=1.75cm]{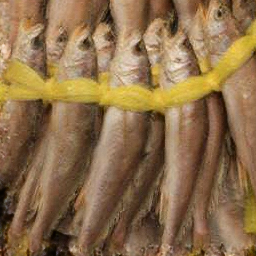}}
			\vspace{-8pt}
			\centerline{\tiny OURS}
		\end{minipage}
	\caption{Qualitative comparison of different methods on the real image dataset collected by Lai \textit{et al.} \cite{lai}.}
 \label{img:detail real dataset}
 \vspace{-0.7cm}
\end{figure}

Several conclusions can be drawn from Fig. \ref{img: ablation experiment}. First, it is difficult for DIP to accurately estimate blur kernels from the limited amount of information, especially when the size of blur kernels is large (the last row in Fig. \ref{img: ablation experiment}). DIP's recovery is over-sharpened, and its estimation of blur kernels is also significantly inaccurate, except when the size of the blur kernel is small (the first row in Fig. \ref{img: ablation experiment}). This is because DIP does not use the variance of pixels to establish constraints as VDIP does, nor does it use gradients to grasp more details as TGV does. Second, the experimental results of VDIP are somewhat unstable due to the absence of TGV constraints. In Fig. \ref{img: ablation experiment}, the results of VDIP in the second and fourth rows are significantly over-sharpened, while the results of VDIP in the first and third rows have some unnatural artifacts. Third, DIP-TGV sometimes has the problem of the sparse MAP. As shown in the third row of Fig. \ref{img: ablation experiment}, DIP-TGV obviously is cursed by the sparse MAP, and the estimated blur kernel was also a delta kernel. When the sparse MAP problem is not generated, DIP-TGV generates similar or slightly better results than DIP, which also indicates that the effect of combining DIP and TGV is not as good as that of combining VDIP and TGV. When VDIP and TGV are combined, the deblurring effect is the best. Finally, regardless the sizes of blur kernels, our method can accurately estimate the blur kernels from images.

\subsection{Comparatitive Experiments}
\begin{table*}\centering
\caption{Comparison of PSNR$\uparrow$ and SSIM$\uparrow$ values  on the synthetic dataset from Lai \textit{et al.} \cite{lai}. }
\begin{tabular}{lcccccc}
\hline
                                 & Manmade     & Natural     & People      & Saturated   & Text        & Average     \\ \hline
Krishnan \textit{et al.} \cite{vdip41}    & 17.70 / 0.463 & 17.69 / 0.465 & 17.62 / 0.460 & 17.51 / 0.452 & 17.66 / 0.458 & 17.64 / 0.459 \\
Pan \textit{et al.} \cite{vdip12}   & 17.84 / 0.483 & 17.57 / 0.484 & 17.23 / 0.476 & 17.10 / 0.473 & 17.49 / 0.501 & 17.45 / 0.483 \\
Kupyn \textit{et al.} \cite{vdip30}   & 17.75 / 0.458 & 17.79 / 0.456 & 18.00 / 0.462 & 18.10 / 0.469 & 18.12 /0.477 & 17.95 / 0.464 \\
Wen \textit{et al.} \cite{vdip67}    & 17.47 / 0.498 & 17.41 / 0.496 & 17.23 / 0.430 & 17.22 / 0.479 & 16.88 / 0.452 & 17.24 / 0.477 \\
Zamir \textit{et al.} \cite{vdip69}    & 17.21 / 0.455 & 17.52 / 0.456 & 17.90 / 0.472 & 18.04 / 0.481 & 17.86 / 0.479 & 17.71 / 0.469 \\
Chen \textit{et al.} \cite{vdip70}   & 16.77 / 0.266 & 18.39 / 0.369 & 19.84 / 0.485 & 18.69 / 0.496 & 17.76 / 0.465 & 18.29 / 0.416 \\
Ren \textit{et al.} DIP\cite{dip37} & 16.08 / 0.280 & 18.85 / 0.407 & 20.64 / 0.476 & 19.83 / 0.508 & 20.34 / 0.547 & 19.15 / 0.444 \\
Dong \textit{et al.} VDIP\cite{vdip} & 16.86 / 0.308 & 19.67 / 0.435 & 22.24 / 0.531 & 21.11 / 0.561 & 21.26 / 0.574 & \underline{20.23} / \underline{0.482} \\
DIP-TGV  & 16.16 / 0.284 & 18.98 / 0.420 & 20.45 / 0.472 & 19.61 / 0.505 & 19.85 / 0.534 & 19.01 / 0.443 \\
OURS                             & 16.94 / 0.315 & 20.15 / 0.466 & 22.64 / 0.559 & 21.40 / 0.578 & 21.51 / 0.595 & \textbf{20.53} / \textbf{0.503} \\ \hline
\end{tabular}
\label{tabel1}
\vspace{-0.3cm}
\end{table*}
\begin{table}\centering
\caption{Comparison of average kernel recovery error$\downarrow$ on the synthetic dataset from Lai \textit{et al.} \cite{lai}. }
\scalebox{0.75}{\begin{tabular}{lcccccc}
\hline
      & Manmade & Natural & People  & Saturated & Text    & Average \\ \hline
Krishnan \textit{et al.} \cite{vdip41} & 0.01559 & 0.01559 & 0.0156  & 0.01562   & 0.01560  & \underline{0.01560}  \\
Pan \textit{et al.} \cite{vdip12} & 0.02601 & 0.02471 & 0.02603 & 0.02218   & 0.02127 & 0.02404 \\
Wen \textit{et al.} \cite{vdip67} & 0.02341 & 0.02613 & 0.02608 & 0.01987   & 0.02130  & 0.02336 \\
Ren \textit{et al.} DIP\cite{dip37}   & 0.01827 & 0.01706 & 0.01688 & 0.01560    & 0.01649 & 0.01686 \\
Dong \textit{et al.} VDIP\cite{vdip}  & 0.02341 & 0.01614 & 0.01543 & 0.01580    & 0.01537 & 0.01723 \\
DIP-TGV   & 0.02170 & 0.01725 & 0.01683 & 0.01614    & 0.01420 & 0.01722 \\
OURS  & 0.01208 & 0.00955 & 0.00907 & 0.01070    & 0.00852 & \textbf{0.00998} \\ \hline
\end{tabular}}
\label{tabel2}
\vspace{-0.3cm}
\end{table}
\begin{figure}
			\begin{minipage}[t]{0.19\linewidth}
				\centering				\centerline{\includegraphics[width=1.75cm,height=1.75cm]{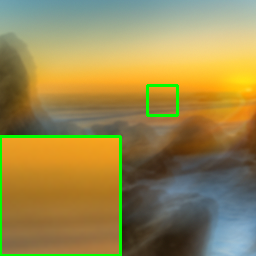}}
    \vspace{-8pt}
				\centerline{\tiny Blurred}
    				\vspace{4pt}
        \centerline{\includegraphics[width=1.75cm,height=1.75cm]{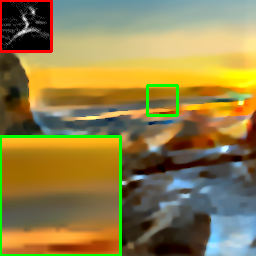}}
        \vspace{-8pt}
				\centerline{\tiny Pan \textit{et al.} \cite{vdip12}}\vspace{4pt}				\centerline{\includegraphics[width=1.75cm,height=1.75cm]{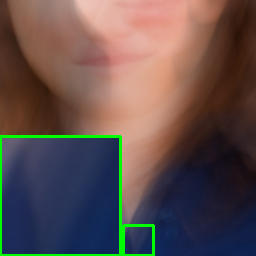}}
    \vspace{-8pt}
				\centerline{\tiny Blurred}
\vspace{4pt}\centerline{\includegraphics[width=1.75cm,height=1.75cm]{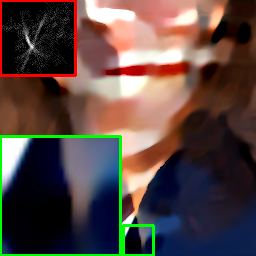}}\vspace{-8pt}
				\centerline{\tiny \cite{vdip12}}\vspace{4pt}				\centerline{\includegraphics[width=1.75cm,height=1.75cm]{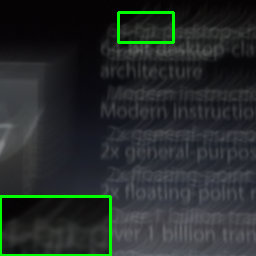}}
    \vspace{-8pt}
				\centerline{\tiny Blurred}\vspace{4pt}
				\centerline{\includegraphics[width=1.75cm,height=1.75cm]{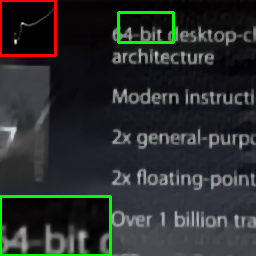}}\vspace{-8pt}
				\centerline{\tiny \cite{vdip12}}
			\end{minipage}
			\begin{minipage}[t]{0.19\linewidth}
				\centering
				\centerline{\includegraphics[width=1.75cm,height=1.75cm]{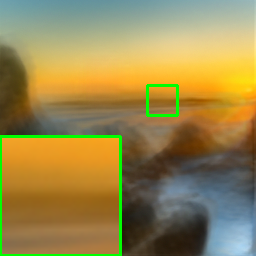}}\vspace{-8pt}				\centerline{\tiny Kupyn \textit{et al.} \cite{vdip30}}\vspace{3pt}
				\centerline{\includegraphics[width=1.75cm,height=1.75cm]{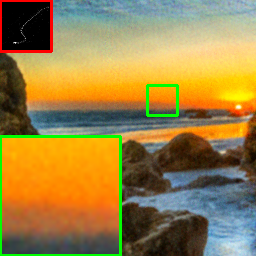}}\vspace{-8pt}
				\centerline{\tiny Ren \textit{et al.} DIP\cite{dip37}}\vspace{4pt}
				\centerline{\includegraphics[width=1.75cm,height=1.75cm]{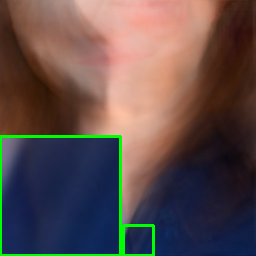}}\vspace{-8pt}
				\centerline{\tiny Kupyn \textit{et al.} \cite{vdip30}}\vspace{3pt}
				\centerline{\includegraphics[width=1.75cm,height=1.75cm]{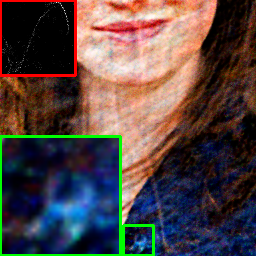}}\vspace{-8pt}
				\centerline{\tiny Ren \textit{et al.} DIP\cite{dip37}}\vspace{4pt}
				\centerline{\includegraphics[width=1.75cm,height=1.75cm]{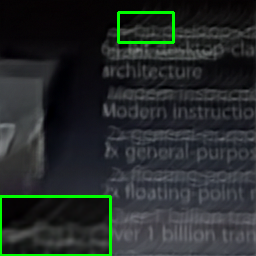}}\vspace{-8pt}
				\centerline{\tiny Kupyn \textit{et al.} \cite{vdip30}}\vspace{2.8pt}	\centerline{\includegraphics[width=1.75cm,height=1.75cm]{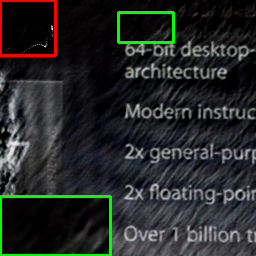}}\vspace{-8pt}
				\centerline{\tiny Ren \textit{et al.} DIP\cite{dip37}}\vspace{4pt}
			\end{minipage}
			\begin{minipage}[t]{0.19\linewidth}
				\centering	\centerline{\includegraphics[width=1.75cm,height=1.75cm]{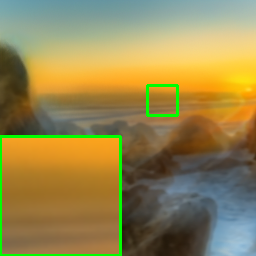}}\vspace{-8pt}
				\centerline{\tiny Zamir \textit{et al.} \cite{vdip69}}				\vspace{3.5pt}\centerline{\includegraphics[width=1.75cm,height=1.75cm]{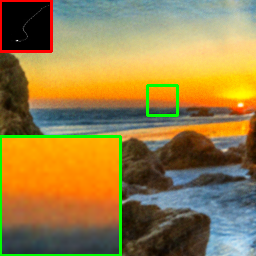}}\vspace{-8pt}
				\centerline{\tiny Dong \textit{et al.} VDIP\cite{vdip}}\vspace{3.45pt}					\centerline{\includegraphics[width=1.75cm,height=1.75cm]{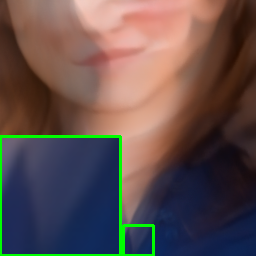}}\vspace{-8pt}
				\centerline{\tiny Zamir \textit{et al.} \cite{vdip69}}\vspace{3.5pt}
\centerline{\includegraphics[width=1.75cm,height=1.75cm]{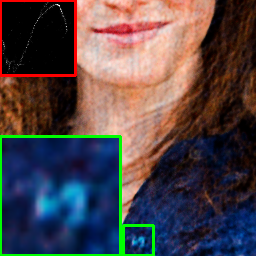}}\vspace{-8pt}
				\centerline{\tiny Dong \textit{et al.} VDIP\cite{vdip}}\vspace{3.5pt}			\centerline{\includegraphics[width=1.75cm,height=1.75cm]{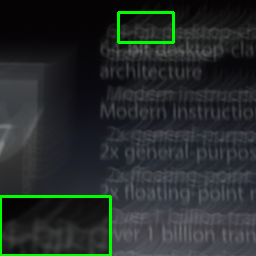}}\vspace{-8pt}
				\centerline{\tiny Zamir \textit{et al.} \cite{vdip69}}
			\vspace{3.2pt}	\centerline{\includegraphics[width=1.75cm,height=1.75cm]{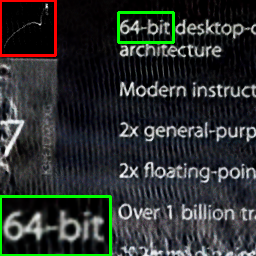}}\vspace{-8pt}
				\centerline{\tiny Dong \textit{et al.} VDIP\cite{vdip}}
			\end{minipage}
			\begin{minipage}[t]{0.19\linewidth}
				\centering
				\centerline{\includegraphics[width=1.75cm,height=1.75cm]{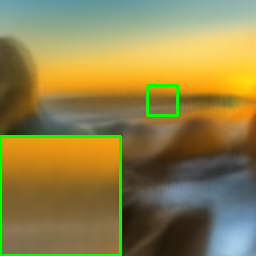}}\vspace{-8pt}
				\centerline{\tiny Chen \textit{et al.} \cite{vdip70} }\vspace{3.5pt}
				\centerline{\includegraphics[width=1.75cm,height=1.75cm]{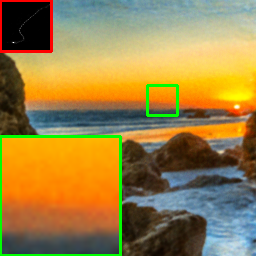}}\vspace{-8pt}
				\centerline{\tiny OURS}\vspace{4.5pt}
				\centerline{\includegraphics[width=1.75cm,height=1.75cm]{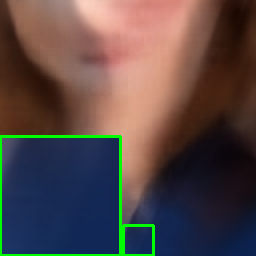}}\vspace{-8pt}
				\centerline{\tiny Chen \textit{et al.} \cite{vdip70} }\vspace{3.5pt}
				\centerline{\includegraphics[width=1.75cm,height=1.75cm]{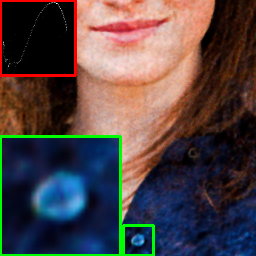}}\vspace{-8pt}
				\centerline{\tiny OURS}\vspace{4.5pt}	\centerline{\includegraphics[width=1.75cm,height=1.75cm]{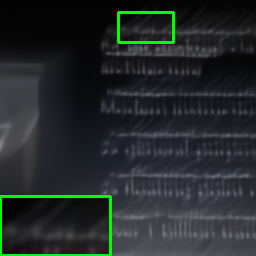}}\vspace{-8pt}
				\centerline{\tiny Chen \textit{et al.} \cite{vdip70} }\vspace{3.2pt}
				\centerline{\includegraphics[width=1.75cm,height=1.75cm]{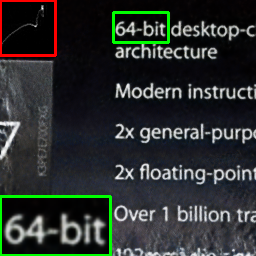}}\vspace{-8pt}
				\centerline{\tiny OURS}
			\end{minipage}
			\begin{minipage}[t]{0.19\linewidth}
				\centering
				\centerline{\includegraphics[width=1.75cm,height=1.75cm]{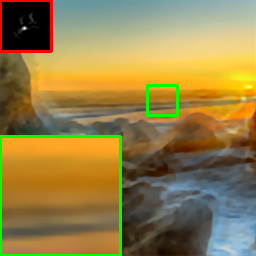}}\vspace{-8pt}
				\centerline{\tiny Wen \textit{et al.} \cite{vdip67}}\vspace{3.5pt}
				\centerline{\includegraphics[width=1.75cm,height=1.75cm]{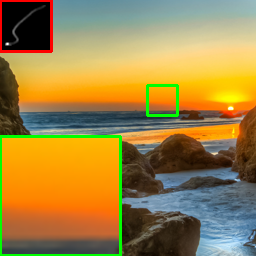}}\vspace{-8pt}
				\centerline{\tiny Clean}\vspace{4.5pt}
				\centerline{\includegraphics[width=1.75cm,height=1.75cm]{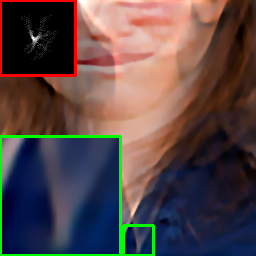}}\vspace{-8pt}
				\centerline{\tiny Wen \textit{et al.} \cite{vdip67}}\vspace{3.5pt}
				\centerline{\includegraphics[width=1.75cm,height=1.75cm]{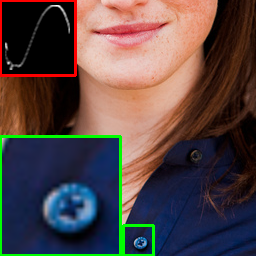}}\vspace{-8pt}
				\centerline{\tiny Clean}\vspace{4.5pt}
	\centerline{\includegraphics[width=1.75cm,height=1.75cm]{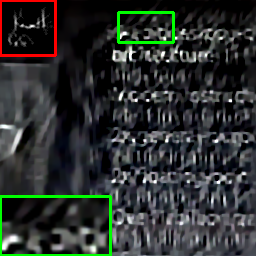}}\vspace{-8pt}
				\centerline{\tiny Wen \textit{et al.} \cite{vdip67}}\vspace{3.2pt}
	\centerline{\includegraphics[width=1.75cm,height=1.75cm]{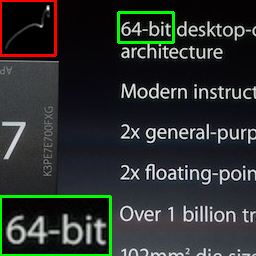}}\vspace{-8pt}
				\centerline{\tiny Clean}
			\end{minipage}		
		\caption{Qualitative comparison on the synthetic dataset from Lai \textit{et al.} \cite{lai}.}
  \label{img: details synthetic dataset}
   \vspace{-0.5cm}
  \end{figure}
\subsubsection{Qualitative Comparison}\label{sec5.3}
To further demonstrate that our method can perform well in visual effects and retain more details, we visualize the recovered images as shown in Figs. \ref{img: details synthetic dataset} and \ref{img:detail real dataset} which are synthetic and real images, respectively. The estimated kernels are put in the upper left of the corresponding deblurring results, and the enlarged details are in the lower left. Since some deep learning-based methods simulate an end-to-end process and do not estimate the kernels, some deblurring results in Fig. \ref{img: details synthetic dataset} do not have kernels shown. 
\vspace{-0.1cm}
Fig. \ref{img: details synthetic dataset} shows the deblurring results of different methods for three synthetic images (nature image, people image, and text image). From Fig. \ref{img: details synthetic dataset}, we can see that our method is visually better and has fewer artifacts than the DIP method. In the deblurring results from traditional methods, many artifacts are over-enhanced and even mismatched. Deep learning-based methods are obviously not effective in deblurring, possibly because there is no sufficient data to train the model. As can be seen from the zoomed details of the natural image, our method yields significantly fewer artifacts and less noise. By examining the people image, it can be seen that the button details are only recovered by our method, thanks to the edge extraction ability of the TGV regularization. From the zoomed details, we observe that characters in the text image are also restored more clearly by our method. 
Inspecting the estimated blur kernels, we can also easily find that our method estimates the blur kernels most accurately.

Observing the deblurring effects on real images in Fig. \ref{img:detail real dataset}, it can be found that deep learning-based methods either have no deblurring effect or generate too smooth images, while the traditional methods are mostly too enhanced. Moreover, the DIP-based methods produce more artifacts than our method. Therefore, we summarize that our approach works best visually, both for synthetic and real images.

\subsubsection{Quantitative Comparison}\label{sec5.4}
Blind deblurring data sets can be divided into three categories: synthetic images, simulated real images,
and real images. 
The difference between simulated 
real images and synthetic images is that the blur kernel of the former is simulated to imitate the real camera shake, while the latter is not. For traditional methods, the well-performing prior to all three kinds of images is difficult to design. For deep learning methods, training on a dataset containing three types of images is very resource-intensive. As a result, there are few methods that work well on three datasets at the same time. The prior information required by our method can directly be obtained from the destroyed single image and the network structure, so our method can be well applied to all images. To prove the above point, we conduct quantitative experiments on all three types of image sets.

For synthetic images, we illustrate the superiority of our method by calculating three numerical indexes: PSNR, SSIM, and MSE. As shown in Tabs. \ref{tabel1} and \ref{tabel2}, the best results are bold-faced, and the second-best are underlined. On the dataset from Lai \textit{et al.} \cite{lai}, our method yields superior image quality. For all three indexes, our model outperforms other competitors. To evaluate the accuracy of the estimated kernels, we calculate the average MSE, and the results are shown in Tab. \ref{tabel2}. Because some end-to-end deep learning models do not estimate blur kernels, Tab. \ref{tabel2} excludes them. It can be seen from Tab. \ref{tabel2} that the blur kernels estimated by our model are the most accurate compared with other models. Additionally, the numerical results in Tab. \ref{tabel1} further demonstrate that only when VDIP and TGV are combined, the deblurring effect is the best. In Tab. \ref{tabel1}, the numerical effect of DIP-TGV is inferior to DIP, which may be related to the sparse MAP problem caused by DIP-TGV.
\begin{table}\centering
\caption{Comparison of PSNR$\uparrow$ and SSIM$\uparrow$  values  on the dataset  from  Kohler \textit{et al.} \cite{ko}. }
\begin{tabular}{lll}
\hline
     & PSNR           & SSIM          \\ \hline
Krishnan \textit{et al.} \cite{vdip41}   & 23.19          & 0.779         \\
Pan \textit{et al.} \cite{vdip12}    & \underline{26.21}          & \underline{0.839}          \\
Kupyn \textit{et al.} \cite{vdip30}   & 25.06          & 0.806          \\
Chen \textit{et al.} \cite{vdip70}   & 25.31          & 0.805          \\
Ren \textit{et al.} DIP\cite{dip37}  & 21.61          & 0.717          \\
Dong \textit{et al.} VDIP\cite{vdip} & 25.33          & 0.850          \\
OURS & \textbf{26.49} & \textbf{0.860} \\ \hline
\end{tabular}
\label{tabelko}
\vspace{-0.2cm}
\end{table}
\begin{table} \centering
\caption{Comparison of NIQE$\downarrow$, BRISQUE$\downarrow$ and PIQE$\downarrow$ values  on the  real blurred dataset  from Lai \textit{et al.} \cite{lai}. }
\begin{tabular}{lcccc}
\hline
 & \multicolumn{1}{c}{NIQE} & \multicolumn{1}{c}{BRISQUE} & \multicolumn{1}{c}{PIQE} & \multicolumn{1}{c}{Average} \\ \hline
Krishnan \textit{et al.} \cite{vdip41}   & 8.8243 & 33.3825 & 51.2052 & 31.1373 \\
Pan \textit{et al.} \cite{vdip12}   & 7.6492 & 30.5786 & 61.6742 & 33.3007 \\
Kupyn \textit{et al.} \cite{vdip30}   & 4.8964 & 32.6875 & 42.5616 & 26.7152 \\
Wen \textit{et al.} \cite{vdip67}   & 8.4819 & 31.6044 & 52.5347 & 30.8737 \\
Zamir \textit{et al.} \cite{vdip69}   & 5.3565 & 36.9692 & 49.4281 & 30.5846 \\
Chen \textit{et al.} \cite{vdip70}   & 6.6433 & 39.4483 & 62.0968 & 36.0628 \\
Ren \textit{et al.} DIP\cite{dip37} & 6.5901 & 30.5180  & 35.4707 & 24.1929 \\
Dong \textit{et al.} VDIP\cite{vdip} & 5.5079 & 28.3879 & 32.7918 & \underline{22.2292} \\
OURS                             & 5.4210  & 29.2051 & 31.8626 & \textbf{22.1629} \\ \hline
\end{tabular}
\label{tabel3}
\vspace{-0.3cm}
\end{table}

For simulated real images, we conduct experiments on a benchmark dataset collected by Kohler \textit{et al.} \cite{ko}. Although this dataset does not have the ground truth like the synthetic images, Kohler \textit{et al.} devised some computational rules to compute PSNR and SSIM similarly. We use the same way as Kohler \textit{et al.} to compute PSNR and SSIM. As can be seen from Tab. \ref{tabelko}, regardless of PSNR or SSIM, the image recovered by our method has the highest performance.

In order to further prove that our method is also applicable to real scenes, we conduct experiments on real images of the dataset (Lai \textit{et al.} \cite{lai}). Since the real image has no ground truth, we utilize three no-reference image quality assessment metrics to quantitatively evaluate the results, as shown in Tab. \ref{tabel3}. It is shown that from the angle of NIQE and PIQE, the image quality after deblurring by our method is the highest, while regarding BRISQUE, the image quality by our method is suboptimal. Overall speaking, our method is still the best.

In brief, our systematic experiments demonstrate that the deep learning-based models do not perform as well as our method, probably because deep learning-based models are highly dependent on a large number of image pairs for training. The traditional method is not as good as ours, either, because the constraints added by the traditional methods are not enough to capture all the information of the images, while our method uses TGV and VDIP, combining the best of two worlds to capture more complete information.

\section{Conclusion}
In this article,  we have proposed a model called VDIP-TGV that can improve VDIP in a highly nontrivial manner by overcoming its two pitfalls. One is that the image will lose details and boundaries after deblurring, and the other is that when the blur kernel size is relatively large, the image will be excessively sharpened after deblurring. Then, we have used the ADMM algorithm to decompose the target problem into four sub-problems to solve. Finally, we have conducted systematic experiments to verify the superiority of our model qualitatively and quantitatively. Future work can be generalizing the proposed VDIP-TGV into other low-level computer vision problems.

\section*{Declaration of competing interest}
	The authors declare that they have no known competing financial interests or personal relationships that could have appeared to influence the work reported in this paper.

\bibliographystyle{IEEEtran}
\bibliography{reference2}

\begin{IEEEbiography}[{\includegraphics[width=1in,height=1.25in,clip,keepaspectratio]{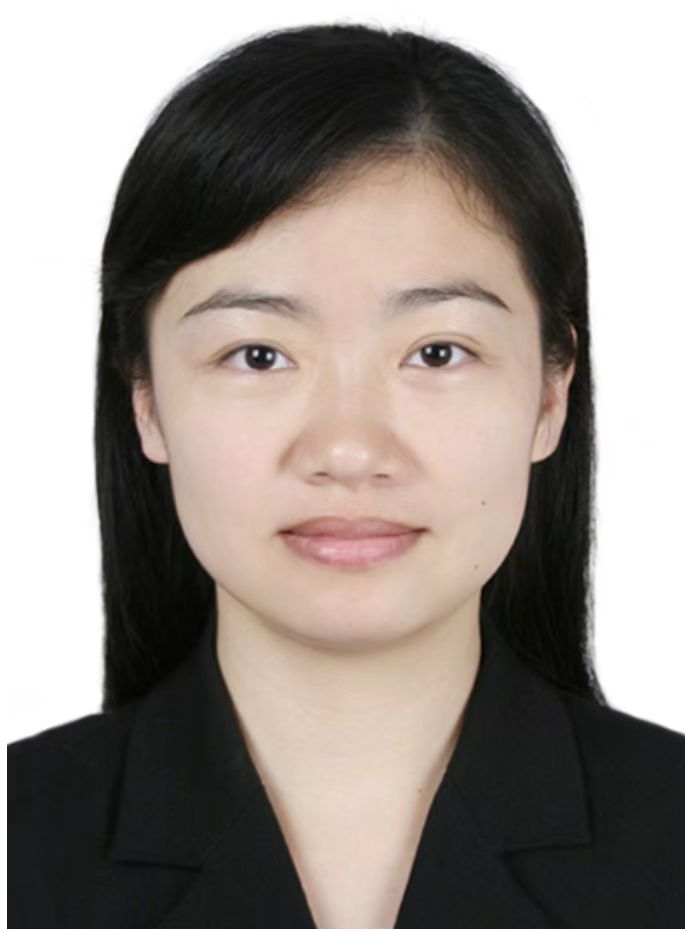}}]{Tingting Wu}  received the B.S. and Ph.D. degrees
 in mathematics from Hunan University, Changsha,
 China, in 2006 and 2011, respectively. From 2015
 to 2018, she was a Postdoctoral Researcher with the
 School of Mathematical Sciences, Nanjing Normal
 University, Nanjing, China. From 2016 to 2017, she
 was a Research Fellow with Nanyang Technological
 University, Singapore. She is currently an Associate
 Professor with the School of Science, Nanjing Uni
versity of Posts and Telecommunications, Nanjing.
 Her research interests include variational methods
 for image processing and computer vision, optimization methods and their
 applications in sparse recovery, and regularized inverse problems.
\end{IEEEbiography}
\begin{IEEEbiography}[{\includegraphics[width=1in,height=1.25in,clip,keepaspectratio]{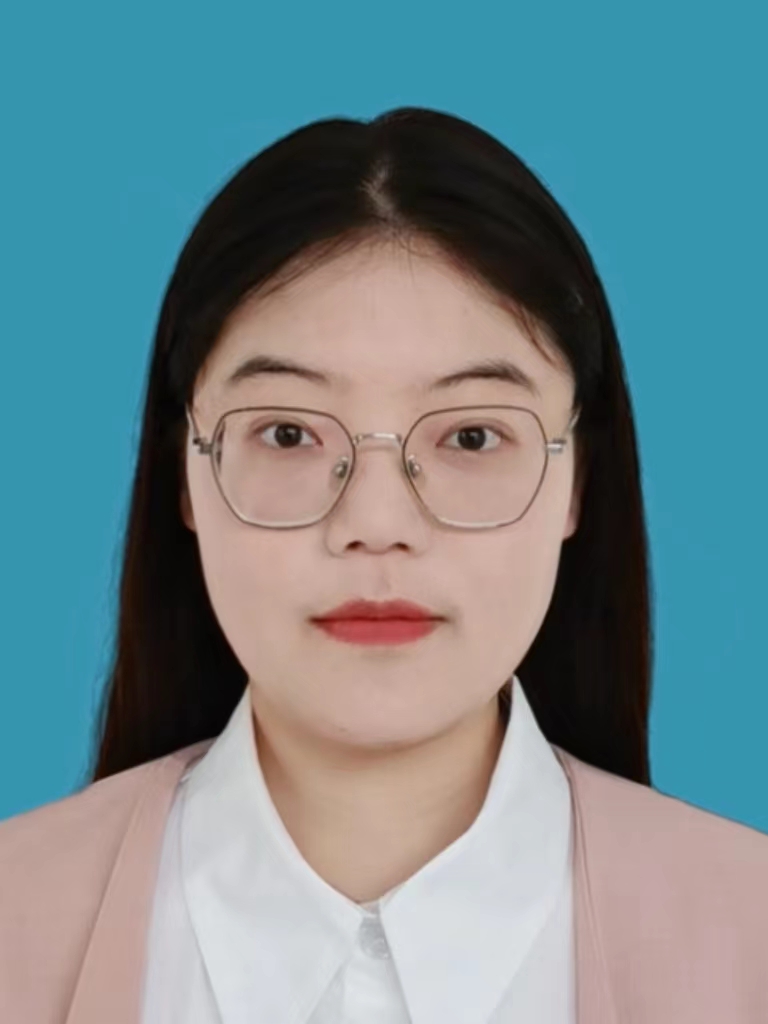}}]{Zhiyan Du} received the B.S. degree in School of Mathematics and Statistics from the Huanghuai University, Zhumadian, China, in 2021.
She is currently pursuing an M.S. degree from the
School of Science, Nanjing University of Posts and
Telecommunications, Nanjing, China. Her research
interests include image processing and computer
vision, machine learning, and inverse problems.
\end{IEEEbiography}
\begin{IEEEbiography}[{\includegraphics[width=1in,height=1.25in,clip,keepaspectratio]{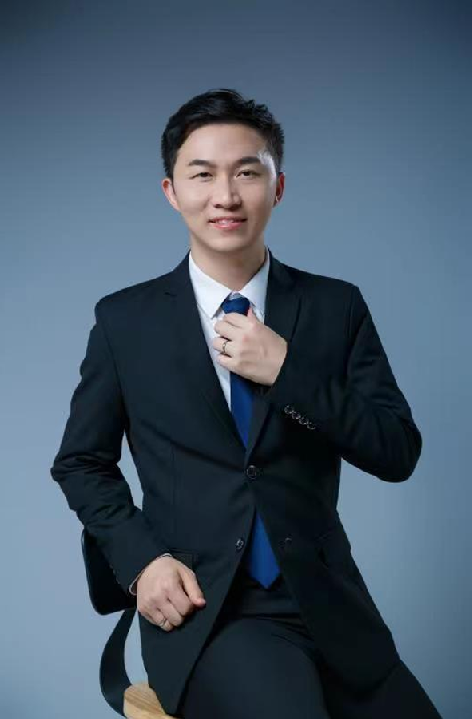}}]{Fenglei Fan} is a research assistant professor at Department of Mathematics, The Chinese
 University of Hong Kong. He received his PhD degree in Rensselaer Polytechnic Institute, US. His research interests lie
 in deep learning theory and methodology. He was the
 recipient of the IBM AI Horizon Fellowship and the
 2021 International Neural Network Society Doctoral
 Dissertation Award.
\end{IEEEbiography}
\begin{IEEEbiography}[{\includegraphics[width=1in,height=1.25in,clip,keepaspectratio]{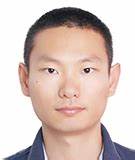}}]{Zhi Li}
obtained his Bachelor's and Master's degrees from the China University of Petroleum, Shandong, China in 2007 and 2010 respectively. He also completed his Master's degree in Applied Science from Saint Mary's University, Halifax, NS, Canada in 2012. He earned his Ph.D. from Hong Kong Baptist University in 2016 and subsequently held a postdoctoral position at Michigan State University, East Lansing, Ml, USA from 2016 to 2019. He currently holds the position of Associate Researcher at the Department of Computer Science and Technology, East China Normal University, Shanghai, China.
\end{IEEEbiography}
\begin{IEEEbiography}[{\includegraphics[width=1in,height=1.25in,clip,keepaspectratio]{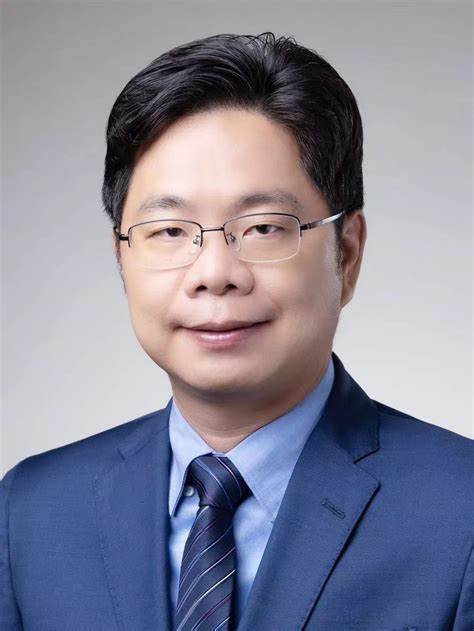}}]{Tieyong Zeng} received the B.S. degree from Peking
University, Beijing, China, in 2000, the M.S degree from Ecole Polytechnique, Palaiseau, France, in 2004, and the Ph.D. degree from the University
of Paris XIII, Paris, France, in 2007. He worked as
a Post-Doctoral Researcher with ENS de Cachan,
Cachan, France, from 2007 to 2008, and an Assistant/Associate Professor with Hong Kong Baptist
University, Hong Kong, from 2008 to 2018. He is
currently a Professor in the Department of Mathematics, The Chinese University of Hong Kong, Hong
Kong. His research interests are image processing, machine learning, and
scientific computing.
\end{IEEEbiography}

\end{document}